%% file: main.tex
\documentclass[11pt]{article}
\usepackage{amsmath}
\usepackage{amssymb}
\usepackage{booktabs}
\usepackage{multirow}
\usepackage{float}
\usepackage{titletoc} 
\usepackage{adjustbox}
\usepackage{tcolorbox}
\definecolor{cream_yellow}{HTML}{FFF8DC}
\definecolor{very_light_yellow}{HTML}{FFF9C4}
\definecolor{pale_yellow}{HTML}{FFF3B0}
\definecolor{soft_amber}{HTML}{FFE8A3}
\usepackage[preprint]{acl}

\usepackage{times}
\usepackage{latexsym}

\usepackage[T1]{fontenc}

\usepackage[utf8]{inputenc}

\usepackage{microtype}

\usepackage{inconsolata}

\usepackage{graphicx}

\usepackage[position=bottom]{subfig}
\usepackage{makecell}
\usepackage{titletoc} 
\definecolor{bg_gray}{RGB}{245,245,245}
\usepackage{enumitem}
\usepackage{geometry}
\usepackage{amsmath}
\usepackage{xcolor}
\usepackage{hyperref}       
\usepackage{url}            
\usepackage{booktabs}       
\usepackage{amsfonts}       
\usepackage{nicefrac}       
\usepackage{microtype}      
\usepackage[table,xcdraw]{xcolor}
\usepackage{enumitem}
\usepackage{colortbl}
\usepackage{threeparttable}
\usepackage{multirow}
\usepackage{pifont} 
\usepackage{graphicx}
\usepackage{amsmath} 
\usepackage{stfloats} 
\usepackage{caption}
\usepackage{subcaption}
\usepackage{booktabs} 
\usepackage[normalem]{ulem}

\title{\raisebox{-0.15cm}{\includegraphics[height=1.3em]{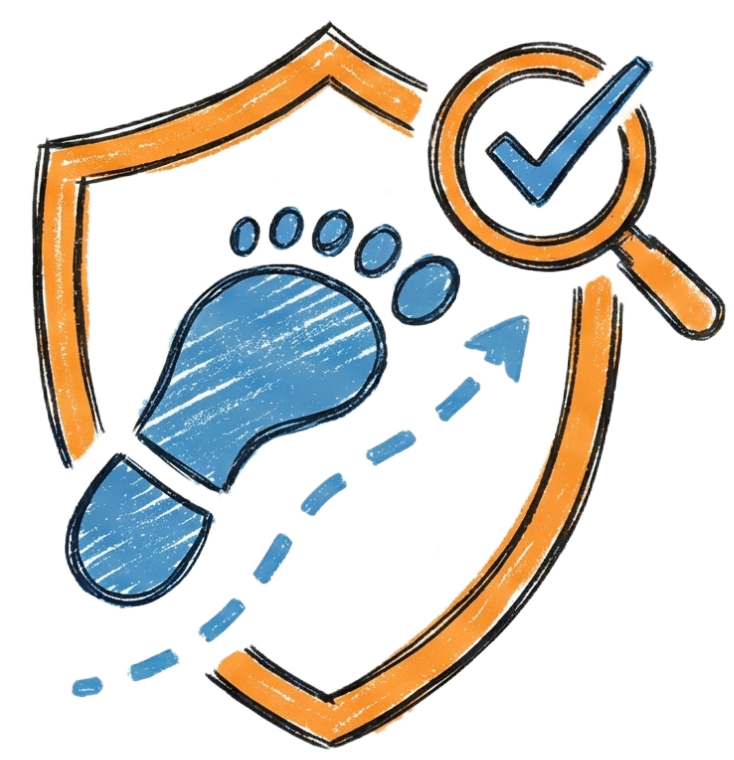}}~ StepGuard: Guarding Web Navigation via Single-Step Calibration}

\author{
\textbf{Zhihao Cui}$^{\textbf{1},\dagger}$ \quad
\textbf{Yuchen Zhang}$^{\textbf{1},\dagger}$ \quad
\textbf{Xiyang Sun}$^{\textbf{3}}$ \quad
\textbf{Yaxiong Wang}$^{\textbf{2*}}$ \quad
\textbf{Li Zhu}$^{\textbf{1}}$ \\
\textbf{Jinpeng Hu}$^{\textbf{2}}$ \quad
\textbf{Liu Liu}$^{\textbf{2}}$ \quad
\textbf{Mengjia Li}$^{\textbf{4}}$ \quad
\textbf{Yujiao Wu}$^{\textbf{5}}$ \\
$^{1}$School of Software Engineering, Xi'an Jiaotong University \\
$^{2}$School of Computer Science and Information Engineering, Hefei University of Technology \\ 
$^{3}$Xiamen University  \quad $^{4}$Zhejiang Lab \quad 
$^{5}$CSIRO \\
{\small $^{\dagger}$Equal contribution.} \quad
{\small $^{*}$Corresponding author.} \\
{\small wangyx@hfut.edu.cn}
}

\begin{document}
\maketitle

\input{abs}
\input{intro}

\input{relatedwork}
\input{prelim}

\input{method}

\input{experim}

\input{conclusion}
\clearpage

\input{limit}

\bibliography{custom}

\clearpage

\input{appendix}

\end{document}

%% file: abs.tex
\begin{abstract}
Web navigation requires agents to follow natural language goals, interact with web pages, and produce accurate answers. While recent advances leverage vision-language models and reinforcement learning, existing methods still suffer from single-step fragility due to reward misalignment and error propagation. To tackle the reward entanglement, we design Dynamic Dual-Policy Optimization (DDPO), which dynamically switches between a navigation-first mode for exploration and an answer-first mode for question-answering to mitigate reward conflict. To calibrate the single-step error, we propose Confidence-Guided Adaptive Navigation Reflection (CANR), a mechanism that estimates per-step confidence, triggers reflection only when necessary, and uses contrastive rewards to encourage self-correction to calibrate the single-step inaccuracy. With the above as the main components, we finally develop our StepGuard, a new framework of Guarding Web Navigation via Single-Step Calibration. Experiments demonstrate that our approach significantly improves navigation and answer accuracy, setting new state-of-the-art performance on standard web navigation benchmarks.
\end{abstract}

%% file: intro.tex
\section{Introduction}
Web navigation aims to enable an intelligent agent to follow a user’s natural language question together with an auxiliary description, perform a sequence of interactions on web pages, and finally reach a target webpage to obtain the required information and generate an answer. This capability is essential for building autonomous web agents that can actively browse, search, and reason over online content. Effective web navigation supports a wide range of real-world applications, including intelligent personal assistants, automated information retrieval, online task execution, and decision support systems, significantly improving how users access information and complete daily tasks on the web. Given its practical significance, extensive efforts have been dedicated to this problem in recent years ~\cite{MiniWoB++, RUSS, FLIN, WebQA, ScreenQA, Synapse, SeeAct, WebArena}.

\begin{figure}[t]
\centering
  \includegraphics[width=\linewidth]{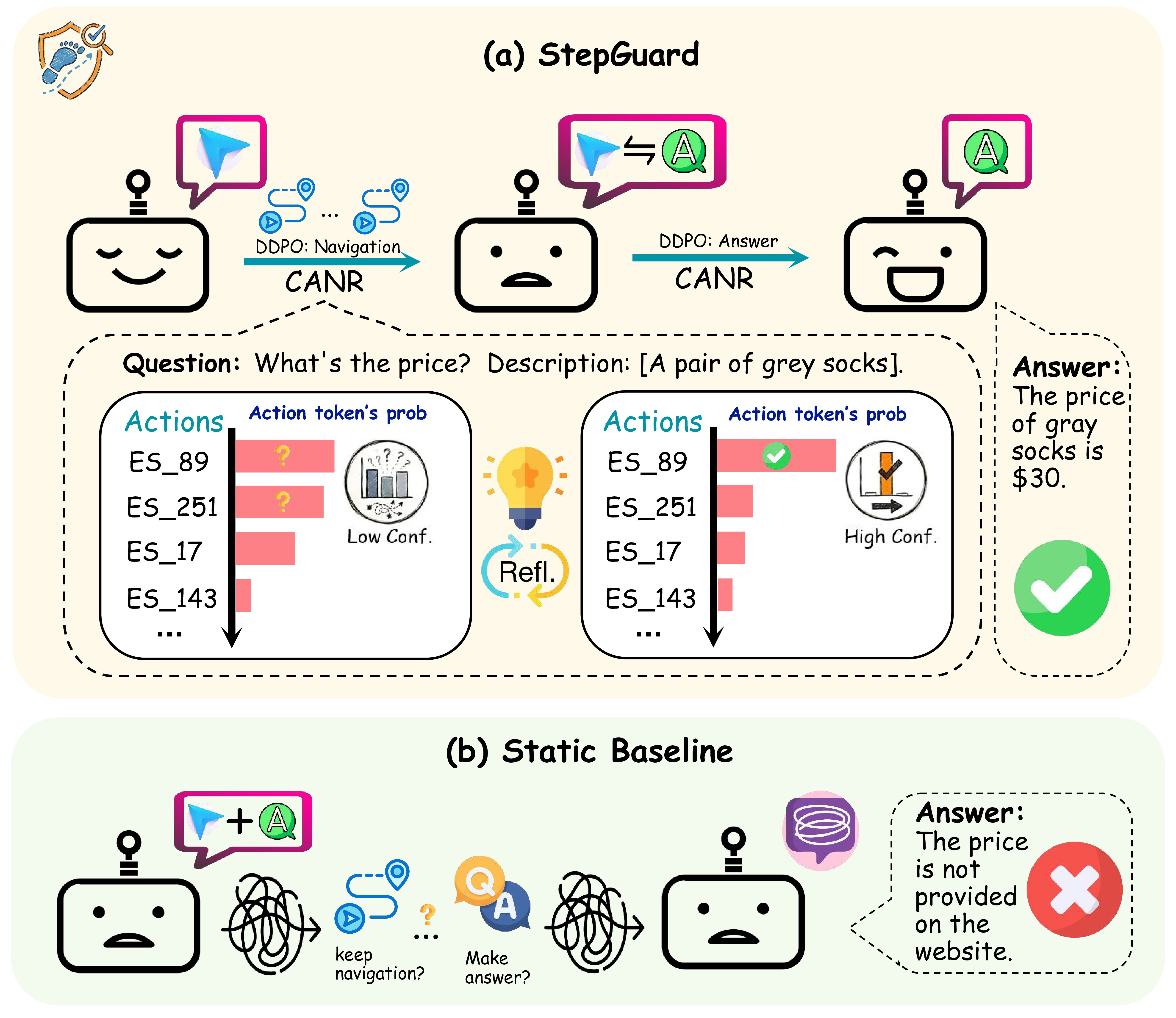}
  \vspace{-0.75cm}
  \caption{
Overview of the proposed StepGuard. Unlike the static baseline (b) that fails due to entangled navigation and answering objectives, StepGuard (a) achieves robust performance by dynamically decoupling these tasks via DDPO and rectifying decisions through CANR, enabling precise state-adaptive execution.
}
\vspace{-0.4cm}

\label{fig:teaser}
\end{figure}

Recently, web navigation has made notable progress driven by advances in Vision-Language Models (VLMs) and Reinforcement Learning (RL). Modern VLMs, such as GPT-4~\citep{GPT4TR}, InternVL~\citep{InternVL}, and Qwen-VL~\cite{qwenvl}, demonstrate strong capabilities in multimodal context understanding, visual question answering, and instruction following, making them effective foundation models for web navigation. Building upon these representations, reinforcement learning further improves the models’ ability to perform sequential decision-making by optimizing action selection through interaction with the web environment~\cite{DigiRL, Agent-FLAN}. With carefully designed reward functions, RL-based training significantly enhances the effectiveness of web agents, leading to consistent performance improvements in navigation and task completion~\cite{AutoWebGLM, WebRL, ShinnCGNY23}.

For web navigation, accurate execution at each step is critical to the success of the overall task~\cite{Mind2Web, WebArena, WebShop}. However, existing methods remain fragile at the single-step level due to two key issues. First, current approaches typically optimize agents using both task-level navigation and question-answering rewards simultaneously, leading to entangled and conflicting optimization objectives~\cite{WebGPT, WebArena}. Navigation rewards encourage prolonged exploration to gather sufficient information, whereas question-answering rewards favor early termination and rapid answer generation~\cite{CPAS, ARAG}. These objectives can conflict across different navigation states, making it difficult for a single policy to balance both effectively~\cite{Multi-Objective-RL-Guide}. To address this issue, we propose Dynamic Dual-Policy Optimization (DDPO), which decouples policy learning into two complementary modes: Navigation-first and Answer-first. The Navigation-first mode trains the agent as an explorer, prioritizing effective information acquisition, while the Answer-first mode treats the agent as a decision-maker, focusing on when to terminate navigation and produce accurate answers. By dynamically switching between the two modes based on the current state, DDPO alleviates reward conflicts and enables more stable policy optimization.

Second, existing web agents are highly vulnerable to single-step decision errors: an incorrect action at one step can propagate through the trajectory, ultimately causing navigation failure and inaccurate answers~\cite{Reflexion, LATS, PlanBench}. To mitigate this issue, we introduce a mechanism that explicitly models uncertainty and calibrates decisions via reflection. Specifically, we propose Confidence-Guided Adaptive Navigation Reflection (CANR), which enhances decision reliability by dynamically reflecting on uncertain steps~\cite{Self-Consistency, Just-Ask-for-Calibration}. CANR first estimates the confidence of the agent’s action at each navigation step. Based on this confidence, a dynamic reflection strategy is employed to selectively trigger reflection only when necessary, avoiding redundant reconsideration. When activated, a contrastive reward encourages the agent to revise its decisions toward more reliable outcomes. Through CANR, the agent acquires the ability to self-monitor, rethink, and refine its actions, resulting in more robust long-horizon navigation behavior.


With DDPO and CANR as core components, we propose StepGuard framework. In summary, we highlight the contributions of this paper as follows:
\begin{itemize}

    \item A \textbf{Confidence-Guided Adaptive Navigation Reflection (CANR)}, which equips the web agent with the ability to self-monitor, selectively reflect, and optimize its action choices, leading to more robust long-horizon navigation behavior.
    
    \item A \textbf{Dynamic Dual-Policy Optimization (DDPO)} strategy that decouples navigation and question-answering objectives through two alternating training modes, thereby stabilizing policy learning and reducing oscillations from conflicting gradients. 

    \item Superior performance on two public datasets, WebVLN and WebWalkerQA. Our StepGuard achieves competitive performance with the state-of-the-art models on both datasets.

\end{itemize}

%% file: relatedwork.tex
\section{Related Work}

\noindent\textbf{Planning and Self-Correction in Web Agents.} To handle dynamic web environments, research has evolved from linear chain-of-thought to structured planning \cite{CoALA}, utilizing verbal memory \cite{Synapse, Expel} or hierarchical state abstractions \cite{WebVoyager, Agent-S, AutoGuide} for long-horizon HTML dependencies. However, robust search-based \cite{LATS} and iterative critiquing frameworks \cite{CRITIC, SelfRefine} often incur high computational costs via static "reflect-always" strategies. In contrast, our Confidence-Guided Adaptive Navigation Reflection (CANR) dynamically triggers reasoning only when uncertainty is detected, effectively optimizing the trade-off between planning rigor and inference efficiency.

\noindent\textbf{Reinforcement Learning for Web Agent Alignment.} While supervised fine-tuning and distillation remain standard for web agents \cite{AgentLumos, GPT4TR, AgentTuning, AutoAct}, they entail significant data and training costs. Reinforcement Learning (RL) provides a data-efficient alternative for grounding \cite{MACPO}. Notably, DeepSeek-Math \cite{DeepSeekMath} introduces Group Relative Policy Optimization (GRPO) to enhance reasoning efficiency without dense value functions. Yet, existing methods like WebRL \cite{WebRL} still struggle with reward entanglement and sparsity in multi-objective scenarios \cite{SPA-RL}. Mitigating this, our StepGuard framework adapts GRPO via Dynamic Dual-Policy Optimization (DDPO), effectively aligning lightweight models by balancing exploration and answer generation.

%% file: prelim.tex
\section{Preliminaries}
Before delving into the details of our method, we first briefly introduce the task of web navigation (WN) and Group Relative Policy Optimization (GRPO) reinforcement learning algorithm.
\subsection{Task Definition}

Given a user instruction consisting of a natural language question and a goal description, WN aims to perform a series of discrete actions (time steps) in the web environment to reach the target page containing the required information and generate the correct answer~\cite{AgentBench, VisualWebArena}.

At time step $t$, the agent receives the current environment state $s_t$, which consists of the following three components: (1) the visual observation corresponding to the current webpage screenshot; (2) the set of candidate actions $C_t$, which includes all interactive elements on the current page; (3) the task instruction provided by the user. The agent selects an action $a_t \in C_t$ based on the current policy $\pi_\theta(a_t \mid s_t)$ and interacts with the web environment to transition to the next state $s_{t+1}$. The action set $C_t$ consists of clickable elements on the page (such as buttons, hyperlinks, and dropdown menu items), and its size changes dynamically with the page structure, making this task characterized by a \textit{large action space} and \textit{non-stationarity}.

When the agent selects the special action $\texttt{[STOP]}$, the navigation phase ends, and the model is required to generate the final answer in natural language form based on the current page context (e.g., a product's price). Thus, the web navigation task is modeled as a finite-horizon Markov Decision Process (MDP), with the objective of maximizing the navigation success rate and answer accuracy within a finite number of steps.

\subsection{Web Navigation via Group Relative Policy Optimization (GRPO)}
To address the limitations of offline imitation learning in open web environments, this paper utilizes the Group Relative Policy Optimization (GRPO) algorithm for optimizing the web navigation agent via online reinforcement learning.

At each training step, the policy $\pi_\theta$ samples $K$ candidate actions from state $s_t$:
\begin{equation} {a_t^1, a_t^2, \dots, a_t^K}, \quad a_t^k \sim \pi_\theta(\cdot \mid s_t). \end{equation}
Each action receives an immediate reward $r_t^k$. GRPO normalizes the rewards and performs relative comparisons within the same sample group to provide an unbiased policy update.
The optimization objective for GRPO is defined as follows:
\begin{equation}
\mathcal{L}_{\text{GRPO}}
=
\mathbb{E}_{a \sim \pi_\theta}
\left[
\sum_{k=1}^{K}
\sigma\left(
\frac{r_t^k - \bar{r}_t}{\tau}
\right)
\log \pi_\theta(a_t^k \mid s_t)
\right],
\label{eq:grpo}
\end{equation}
where $\bar{r}_t$ represents the average reward of the samples in the same group:
\begin{equation}
\bar{r}_t = \frac{1}{K}\sum_{k=1}^{K} r_t^k,
\end{equation}

%% file: method.tex
\section{Methodology}
\begin{figure*}[t]

\centering
  \includegraphics[width=0.96\linewidth]{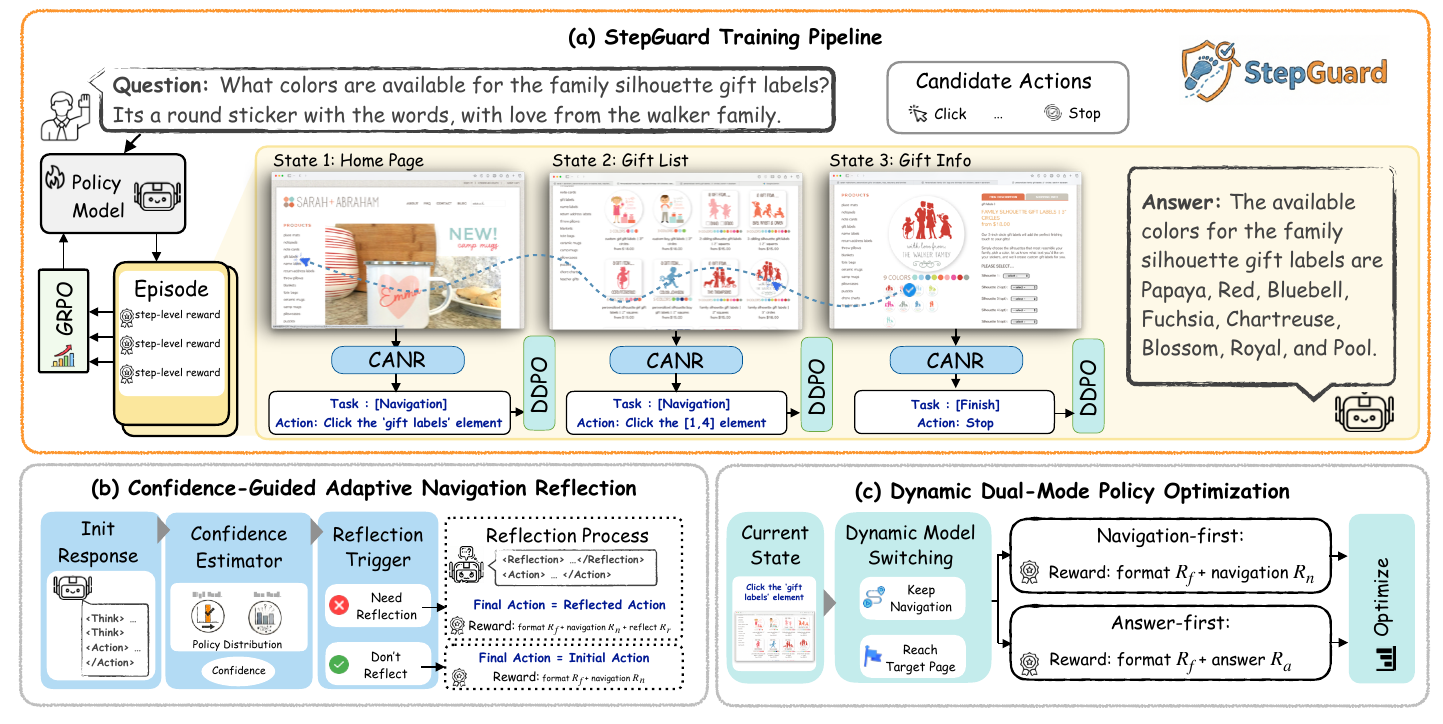}
  \vspace{-0.3cm}
  \caption{Overview of our StepGuard. At each navigation step, CANR estimates the confidence of the action determination and adaptively triggers reflection to suppress single-step errors. Meanwhile, DDPO disentangles step-wise rewards through dynamic switching between navigation-first and answer-first modes. Finally, the entire framework is optimized using the GRPO algorithm. }
  \vspace{-0.4cm}
  
\label{fig:model_arch}
\end{figure*}


\subsection{Overview}

Figure \ref{fig:model_arch} illustrates the overall framework of StepGuard. 
Specifically, StepGuard is designed as a general-purpose framework compatible with both Multimodal and Unimodal Large Language Models, demonstrating high versatility. 
In each state, the environment observation
is processed by the corresponding backbone for representation acquisition. 
Subsequently, the model generates intermediate reasoning steps followed by the action distributions. 
Finally, the carefully crafted rewards from Dynamic Dual-Policy Training and Confidence-Guided Adaptive Navigation Reflection work together with the GRPO algorithm to optimize the overall network. 
In the following, we elaborate on the Dynamic Dual-Policy Training and Confidence-Guided Adaptive Navigation Reflection, which are the key components of our framework.

\subsection{Dynamic Dual-Policy Optimization}

Dynamic Dual-Policy Optimization (DDPO) aims to decouple the entangled rewards of web navigation and question-answering tasks at the single-step level. In web navigation, navigation rewards favor continued exploration, whereas question-answering rewards emphasize timely termination and accurate response generation. Optimizing these heterogeneous objectives jointly can therefore introduce gradient interference~\cite{YuK0LHF20,SenerK18,Multi-Objective-RL-Guide}. DDPO addresses this issue by partitioning single-step interaction data into navigation and question-answering subsets, and alternately optimizing the policy under a navigation-first mode and an answer-first mode.


\noindent\textbf{Navigation-first mode} focuses on guiding the agent toward the target page and incorporates two reward terms. Specifically, navigation reward $R_n$ encourages the agent to take actions that move it closer to the target page, while the format reward $R_f$ ensures that the model outputs are valid and interpretable.

Specifically, the navigation reward $R_n$ encourages the agent to approach the target page $p_\text{target}$ and is computed as:
\begin{equation}
R_n = \begin{cases} 
1 & \text{if } a_t = a_\text{expert} \\
1, &  \text{if } d(s_t, p_\text{target}) < d(s_{t-1}, p_\text{target}) \\
0 & \text{otherwise}
\end{cases}
\label{eq:reward_navigation}
\end{equation}
where $a_t$ represents the current action, $a_\text{expert}$ denotes the expert action (ground-truth annotation), and $s_t, s_{t-1}$ represent the current and previous states, respectively. To calculate the distance $d(s, p)$, we model the website as a directed graph $\mathcal{G}=(V, E)$, where pages $V$ act as nodes and hyperlinks $E$ as edges. The metric $d(s, p)$ is defined as the shortest path length (geodesic distance) from state $s$ to the target page $p$ within this topological structure.

The format reward $R_f$ serves as a fundamental constraint to ensure the agent's output can be parsed by the environment. It is defined as an indicator function based on regular expression matching, mathematically:
\begin{equation}
R_f = \mathbb{I}\left[ \text{Parse}(a_t) \neq \emptyset \right]
\label{eq:reward_format}
\end{equation}
where $\mathbb{I}[\cdot]$ is the indicator function, and $\text{Parse}(\cdot)$ verifies if the output strictly follows the required template: enclosing the thought process in \texttt{<think>} and the action in \texttt{<answer>} tokens.

\noindent\textbf{Answer-first mode} solely aims to generate accurate answers while paying no attention to the navigation and only takes the question-answering reward and the format reward.

Particularly, we compute the question-answering reward $R_a$ based on the F1 score between the generated answer and the ground truth answer.
\begin{equation}
R_a = F1(\hat{y}, y_\text{gt})
\label{eq:reward_question_answering}
\end{equation}
where $\hat{y}$ is the model's predicted answer and $y_\text{gt}$ is the ground truth. The format reward is with the same definition as Eq.\ref{eq:reward_format}.

\noindent\textbf{Alternating Dual-Mode Optimization.} Let $z_t$ denote the task type of a single-step training sample. DDPO does not rely on online oracle signals to switch modes at inference time. Instead, during training, it alternates between the navigation subset and the question-answering subset, applying one non-conflicting objective at a time:
\begin{equation}
R_{DDPO} = \begin{cases}
R_a + R_f, & \text{if } z_t \in \mathcal{D}_{qa},  \\
R_n + R_f, & \text{if } z_t \in \mathcal{D}_{nav}.
\end{cases}
\label{eq:reward_total}
\end{equation}
where $\mathcal{D}_{qa}$ and $\mathcal{D}_{nav}$ denote the pre-partitioned question-answering and navigation subsets, respectively. This data- and reward-level decoupling strips heterogeneous supervision at the source, helps mitigate conflicting gradients, and yields more stable policy updates. During inference, the agent relies solely on its learned policy without any oracle annotation or online mode label.

DDPO utilizes step-wise rewards and also remediates the sparse rewards for task-level rewards mechanisms~\cite{SPA-RL}.

\subsection{Confidence-Guided Adaptive Navigation Reflection}
To remedy the overall task failure caused by one-step error, CANR first estimates the confidence and performs a dynamic reflection for calibration.  

\noindent\textbf{Navigation Confidence Estimation.} We first measure the degree of certainty the model has in its action selection at a given state. Specifically, given state $s_t$ and a set of candidate actions $C_t$, the navigation confidence is defined as the KL divergence between the policy distribution $\pi_\theta(\cdot \mid s_t)$ and the uniform distribution $U(C_t)$:
\begin{equation}
Conf(s_t) = D_{KL}\big(\pi_\theta(\cdot \mid s_t) \,\|\, U(C_t)\big).
\label{eq:conf}
\end{equation}
A larger value of $Conf(s_t)$ indicates that the model has a clear preference for a particular action, while a value close to zero suggests uncertainty between multiple options. Under a fixed action space, this score is equivalent to negative entropy up to a constant shift.\footnote{See the \hyperref[supp:sec.B]{Appendix B}.} We therefore use it as a relative uncertainty signal rather than a direct probability of correctness.

\noindent\textbf{Probabilistic Reflection Trigger.} With the estimated confidence, we  perform an adaptive reflection accordingly. 
In particular, we normalize the confidence score as the probability of triggering reflection:
\begin{equation}
P_{\text{reflect}}(s_t) = \exp\big(-\kappa \cdot Conf(s_t)\big),
\label{eq:preflect}
\end{equation}
where $\kappa$ is a scaling hyperparameter controlling the sensitivity of the reflection trigger to confidence and is empirically set as 4  in our experiments. We trigger the reflection with probability of $P_{\text{reflect}}$,
ensuring that reflection is more likely when the model’s confidence is low, reducing unnecessary reasoning when the decision is clear.

 \noindent\textbf{Calibration via Successful Reflection Reward}. When reflection is activated, the model follows a two-stage generation process. In the first stage, the model generates an initial response $y_{init}$ under the current input conditions, which includes reasoning steps denoted as \texttt{<think><think>} and the final decision as \texttt{<answer><answer>}. In the second stage, predefined reflection prompts\footnote{{See the \hyperref[supp:sec.A]{Appendix A}.}} are inserted to guide the model in re-evaluating its previous decision, generating a revised response $y_{ref}$. The format of $y_{ref}$ is identical to that of $y_{init}$. This process can be seen as a lightweight metacognitive intervention, using explicit prompts to encourage the model to reassess whether its action choices are consistent with the current page's evidence.

The final output is formed by concatenating these parts:
\begin{equation}
y_{\text{final}} = y_{init} \oplus \text{reflection} \oplus y_{ref},
\label{eq:yfinal}
\end{equation}
where $\oplus$ denotes sequence concatenation.

\begin{figure*}[t]

\centering
  \includegraphics[width=0.96\linewidth]{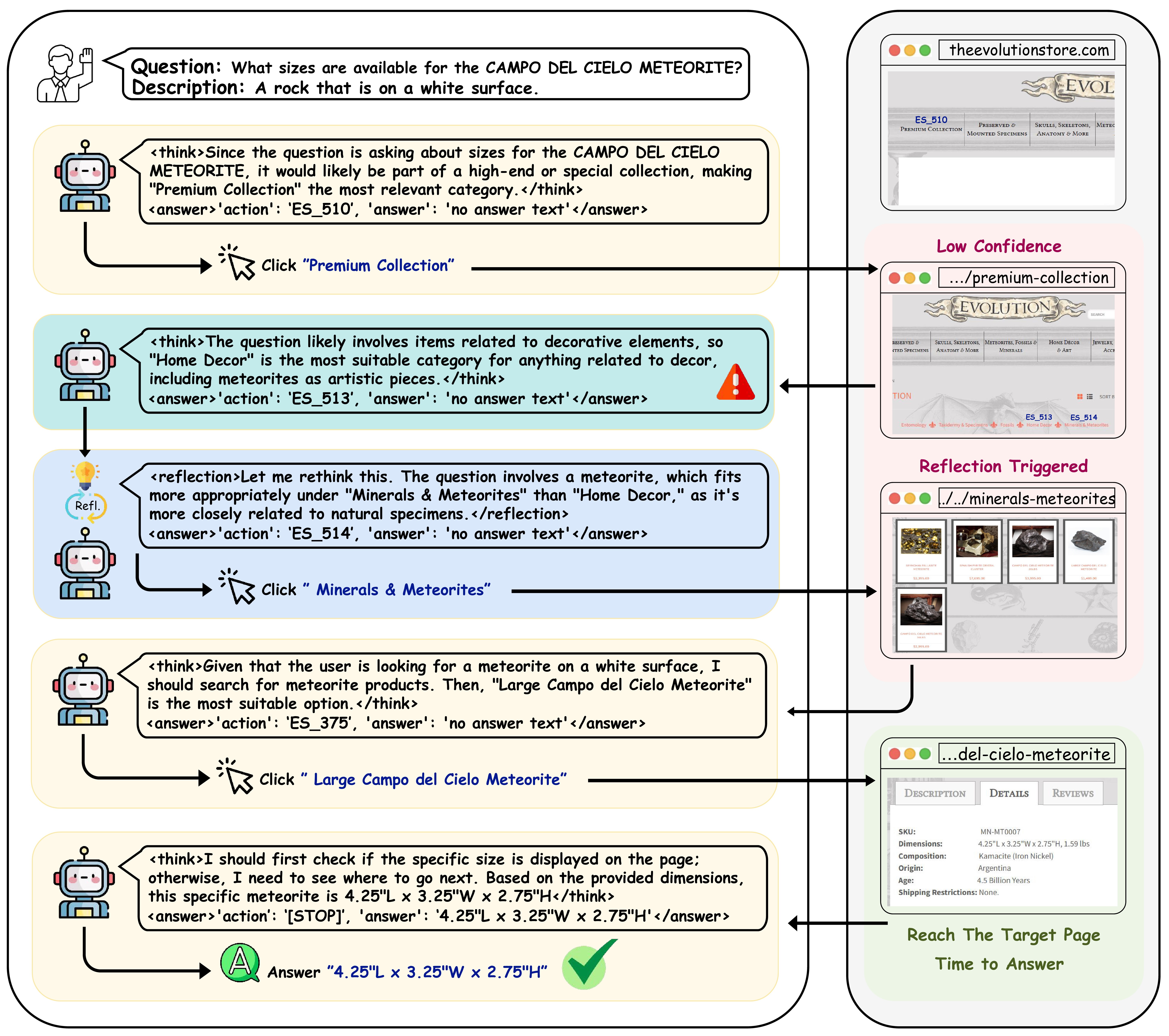}
  \vspace{-0.3cm}
  \caption{A qualitative example of StepGuard's reasoning trajectory, illustrating the self-correction process triggered by adaptive reflection. StepGuard can adaptively perform the reflection if an uncertain action is determined, offering more reliable single-step action decisions for web navigation.}
  \vspace{-0.35cm}
  
\label{fig:example}
\end{figure*}

To ensure effective correction after reflection, we introduce a \textit{reflection success reward} $R_{\text{rs}}$ on top of the original reward function. Since not all reflections contribute to performance improvement, we only encourage \emph{successful reflections}, defined as those that yield higher rewards than their previous outputs. With this consideration, we design the successful reflection rewards. Specifically, let $a_t^{init}$ represent the action parsed from the initial response $y_{init}$, and let $a_t^{ref}$ represent the action parsed from the reflection response $y_{ref}$. The navigation reward after reflection is defined as:
\begin{equation}
R_{\text{rs}}^{\text{nav}} = \mathbb{I}\big[ R_n(s_t,a_t^{init}) < R_n(s_t,a_t^{ref}) \big], 
\label{eq:rs_nav}
\end{equation}
where $\mathbb{I}[\cdot]$ is the indicator function. 
We further introduce the format reward and navigation reward to form the final reward for the reflection: 
\begin{equation}
 R_{ref}(s_t,a_t) = R_f + R_n + R_{\text{rs}}.
\end{equation}

In summary, the training process of CANR reinforces the behavior chain of "low confidence $\rightarrow$ trigger reflection $\rightarrow$ correct decision", improving the model's robustness and efficiency during online policy learning.

%% file: experim.tex
\section{Experiment}
We evaluate our method on two public benchmarks: WebVLN~\cite{WebVLN} and WebWalkerQA~\cite{WebWalker}. For both datasets, we adhere to established evaluation protocols and report standard performance metrics to ensure a fair comparison with prior works. Detailed descriptions regarding the implementation, dataset statistics, and metric definitions are provided in \hyperref[supp:sec.A]{Appendix A}.

\begin{table*}[t]
\caption{Evaluation results on the WebVLN dataset~\cite{WebVLN}. We report performance on both Validation (Val) and Test splits. VLN$\circlearrowright$BERT~\cite{VLNBERT} and VLN$\circlearrowright$BERT* denote random and LXMERT~\cite{LXMERT} initialization, respectively. WebGUM and WebGUM$^\dagger$~\cite{WebGUM} refer to models based on T5-small and T5-base, respectively.}
\vspace{-0.35cm}
\centering
\resizebox{\textwidth}{!}{
    \begin{tabular}{lcccccccccccc}
    \toprule
    \multirow{2}{*}{Method} & \multicolumn{6}{c}{Val} & \multicolumn{6}{c}{Test} \\
    \cmidrule(lr){2-7} \cmidrule(lr){8-13}
     & SR $\uparrow$ & OSR $\uparrow$ & SPL $\uparrow$ & TL $\downarrow$ & WUPS0.9 $\uparrow$ & WUPS0.0 $\uparrow$
     & SR $\uparrow$ & OSR $\uparrow$ & SPL $\uparrow$ & TL $\downarrow$ & WUPS0.9 $\uparrow$ & WUPS0.0 $\uparrow$ \\
    \midrule
    Random & 0.05 & 0.17 & 0.02 & 6.49 & 0.00 & 0.00 & 0.04 & 0.17 & 0.02 & 6.50 & 0.00 & 0.00 \\
    \midrule
    VLN$\circlearrowright$BERT~\cite{VLNBERT} & 17.59 & 17.59 & 16.73 & 6.81 & 9.99 & 13.91 & 11.28 & 12.04 & 10.75 & 7.55 & 7.12 & 9.26 \\
    VLN$\circlearrowright$BERT*~\cite{VLNBERT} & 18.62 & 18.62 & 18.14 & 6.96 & 11.23 & 14.98 & 12.23 & 12.23 & 11.74 & 7.72 & 8.50 & 10.36 \\
    WebGUM~\cite{WebGUM} & 6.02 & 6.02 & 6.02 & 2.99 & 1.84 & 4.08 & 9.71 & 9.71 & 9.71 & 3.15 & 3.57 & 6.98 \\
    WebGUM$^\dag$~\cite{WebGUM} & 31.22 & 31.78 & 31.22 & 3.44 & 18.26 & 24.88 & 29.29 & 29.39 & 29.26 & 3.44 & 17.34 & 23.48 \\
    WebVLN-Net~\cite{WebVLN} & 39.46 & 39.54 & 39.46 & 3.71 & 24.26 & 31.87 & 34.76 & 34.80 & 34.59 & 4.34 & 22.13 & 28.58 \\
    AgentBench~\cite{AgentBench} & 6.85 & 11.73 & 4.05 & 5.37 & 1.60 & 4.27 & 6.97 & 11.94 & 4.13 & 5.39 & 1.64 & 4.65 \\
    NavGPT~\cite{zhou2024navgpt} & 16.78 & 21.63 & 11.45 & 5.40 & 5.89 & 11.92 & 16.92 & 21.89 & 11.61 & 5.43 & 5.97 & 12.06 \\
    \midrule
    Baseline & 39.61 & 39.90 & 39.24 & 4.65 & 29.58 & 33.15 & 36.72 & 36.80 & 35.98 & 4.38 & 27.85 & 32.12 \\
    \; +DDPO & 41.64 & 41.73 & 41.05 & 4.69 & 30.12 & 35.89 & 38.58 & 38.56 & 37.99 & 4.41 & 29.97 & 33.31 \\
    \; +CANR & 41.41 & 41.96 & 41.22 & 5.26 & 29.98 & 35.97 & 38.36 & 38.42 & 38.15 & 4.94 & 29.13 & 32.50 \\
    \rowcolor{cream_yellow}
    \textbf{\; +DDPO+CANR} & \textbf{42.32} & \textbf{42.41} & \textbf{41.76} & 4.99 & \textbf{30.63} & \textbf{35.92} & \textbf{39.83} & \textbf{39.94} & \textbf{39.21} & 4.58 & \textbf{31.28} & \textbf{33.82} \\
    \bottomrule
    \end{tabular}
}
\vspace{-0.55cm}
\label{tab:results_webvln}
\end{table*}

\subsection{Results}

\noindent\textbf{Quantitative Results.}
Table \ref{tab:results_webvln} and \ref{tab:baseline_comparison} present the quantitative evaluation on WebVLN and WebWalkerQA, where our baseline is QwenVL-2.5-3B trained with GRPO.
On WebVLN,  StepGuard (Baseline+DDPO+CANR) achieves a state-of-the-art Success Rate (SR) of 39.83\%, surpassing the strong baseline WebVLN-Net by a margin of \emph{5.07\%}. The decomposed results further show that DDPO provides the primary gain over the baseline, while CANR alone also consistently improves the baseline and remains complementary when combined with DDPO.
On WebWalkerQA, our method demonstrates exceptional parameter efficiency. Despite utilizing a lightweight 3B backbone, StepGuard achieves 54.25\%, 47.99\%, and 25.38\% SR on Easy, Medium, and Hard subsets, respectively.
Crucially, our 3B model outperforms significantly larger baselines, including Qwen-2.5-14B (23.33\% on Hard) and Qwen-2.5-32B (23.33\% on Hard).
Remarkably, it achieves near-parity with the 72B-parameter model (25.83\% on Hard), suggesting that calibrating single-step reasoning effectively bridges the capability gap between small and large language models in complex navigation tasks.

\begin{table}[!b]
\vspace{-0.49cm}
\caption{Comparisons on the WebWalker dataset~\cite{WebWalker}. We report Success Rate (SR) and Action Count (A.C.).}
\vspace{-0.3cm}

\centering
\setlength{\tabcolsep}{2.5pt}
\resizebox{\columnwidth}{!}{
    \begin{tabular}{lcccccc}
    \toprule
    \multirow{2}{*}{\textbf{Method}} & \multicolumn{2}{c}{\textbf{Easy}} & \multicolumn{2}{c}{\textbf{Medium}} & \multicolumn{2}{c}{\textbf{Hard}} \\
    \cmidrule(lr){2-3} \cmidrule(lr){4-5} \cmidrule(lr){6-7}
     & SR & A.C. & SR & A.C. & SR & A.C. \\
    \midrule

    \multicolumn{7}{l}{\textit{\textbf{GPT-4o}}} \\
    \midrule[0.2pt]
    ReAct~\cite{React}     & 53.75 & 2.53 & 45.00 & 3.34 & 30.00 & 5.61 \\
    Reflexion~\cite{Reflexion} & 56.25 & 2.91 & 51.43 & 3.88 & 30.00 & 5.75 \\
    WebWalker~\cite{WebWalker} & 55.00 & 2.97 & 50.00 & 3.43 & 30.00 & 6.02 \\
    \midrule

    \multicolumn{7}{l}{\textit{\textbf{Qwen-Plus}}} \\
    \midrule[0.2pt]
    ReAct~\cite{React}      & 48.75 & 1.67 & 48.57 & 2.69 & 28.33 & 4.00 \\
    Reflexion~\cite{Reflexion} & 53.75 & 3.66 & 40.00 & 3.79 & 24.17 & 5.88 \\
    WebWalker~\cite{WebWalker} & 55.00 & 3.72 & 47.14 & 3.19 & 30.00 & 6.13 \\
    \midrule

    \multicolumn{7}{l}{\textit{\textbf{Qwen-2.5-7B}}} \\
    \midrule[0.2pt]
    ReAct~\cite{React}      & 37.50 & 3.36 & 18.57 & 4.88 & 9.17 & 5.45 \\
    Reflexion~\cite{Reflexion} & 37.50 & 4.03 & 25.00 & 3.48 & 11.67 & 4.57 \\
    WebWalker~\cite{WebWalker} & 41.25 & 3.39 & 24.71 & 3.86 & 12.50 & 5.93 \\
    \midrule

    \multicolumn{7}{l}{\textit{\textbf{Qwen-2.5-14B}}} \\
    \midrule[0.2pt]
    ReAct~\cite{React}      & 36.25 & 1.86 & 32.14 & 2.75 & 15.00 & 3.61 \\
    Reflexion~\cite{Reflexion} & 46.25 & 2.21 & 34.29 & 2.83 & 15.00 & 4.14 \\
    WebWalker~\cite{WebWalker} & 41.25 & 2.42 & 41.43 & 3.46 & 23.33 & 4.42 \\
    \midrule

    \multicolumn{7}{l}{\textit{\textbf{Qwen-2.5-32B}}} \\
    \midrule[0.2pt]
    ReAct~\cite{React}      & 47.50 & 2.21 & 35.71 & 2.30 & 16.67 & 3.75 \\
    Reflexion~\cite{Reflexion} & 42.50 & 2.52 & 32.86 & 2.65 & 14.33 & 3.64 \\
    WebWalker~\cite{WebWalker} & 41.25 & 2.69 & 34.29 & 3.09 & 23.33 & 4.54 \\
    \midrule

    \multicolumn{7}{l}{\textit{\textbf{Qwen-2.5-72B}}} \\
    \midrule[0.2pt]
    ReAct~\cite{React}      & 47.50 & 1.68 & 38.57 & 2.79 & 20.00 & 3.40 \\
    Reflexion~\cite{Reflexion} & 57.50 & 3.04 & 44.29 & 3.88 & 21.67 & 4.25 \\
    WebWalker~\cite{WebWalker} & 58.75 & 2.70 & 48.57 & 3.07 & 25.83 & 5.71 \\
    \midrule

    \multicolumn{7}{l}{\textit{\textbf{Qwen3B}}} \\
    \midrule[0.2pt]
    ReAct~\cite{React}     & 32.50 & 3.52 & 15.71 & 5.12 & 6.67 & 5.88 \\
    Reflexion~\cite{Reflexion} & 33.75 & 4.21 & 21.43 & 3.65 & 8.33 & 5.20 \\
    WebWalker~\cite{WebWalker} & 36.25 & 3.65 & 20.00 & 4.10 & 10.00 & 6.15 \\
    \midrule
    \multicolumn{7}{l}{\textit{\textbf{Qwen3B(ours)}}} \\
    \midrule[0.2pt]
    Baseline & 47.25 & 3.02 & 40.43 & 3.55 & 19.67 & 6.10 \\
    \; +DDPO & 52.50 & 2.85 & 45.71 & 3.28 & 23.17 & 5.71 \\
    \; +CANR & 51.81 & 3.86 & 44.20 & 4.49 & 23.65 & 6.73 \\
    \rowcolor{cream_yellow}
\textbf{\; +DDPO+CANR} & \textbf{54.25} & 3.78 & \textbf{47.99} & 4.20 & \textbf{25.38} & 6.55 \\
    \bottomrule
    \end{tabular}
}
\vspace{-0.2cm}

\label{tab:baseline_comparison}
\end{table}

\begin{table}[t]
\caption{Ablation study on Step-wise Action Accuracy. This metric measures the probability of the agent selecting the correct action or valid link at each step. Both CANR and DDPO enhance the accuracy.}
\label{tab:step_wise_accuracy}
\vspace{-0.2cm}

\centering
\small 
\setlength{\tabcolsep}{12pt} 

\begin{adjustbox}{max width=\linewidth}
    \begin{tabular}{lcc}
    \toprule
    \textbf{Method} & \textbf{Step Acc (\%)} & \textbf{Improv.} \\
    \midrule
    
    \multicolumn{3}{l}{\textit{\textbf{Dataset: WebVLN}}} \\ 
    \hspace{1em} Baseline & 78.54 & - \\ 
    \hspace{1.5em} + DDPO & 81.23 & +2.69 \\
    \hspace{1.5em} \textbf{+ DDPO + CANR} & \textbf{82.15} & \textbf{+3.61} \\
    
    \midrule 
    
    \multicolumn{3}{l}{\textit{\textbf{Dataset: WebWalkerQA}}} \\
    \hspace{1em} Baseline & 82.10 & - \\
    \hspace{1.5em} + DDPO & 85.65 & +3.55 \\
    \hspace{1.5em} \textbf{+ DDPO + CANR} & \textbf{86.92} & \textbf{+4.82} \\
    
    \bottomrule
    \end{tabular}
\end{adjustbox}
\vspace{-0.5cm}
\end{table}

\noindent\textbf{Step-wise Action Accuracy.}
To further validate the reliability of our approach at a fine-grained level, we report the Step-wise Action Accuracy in Table \ref{tab:step_wise_accuracy}. The results demonstrate that our StepGuard achieves a substantial accuracy improvement of +3.61\% on WebVLN and \emph{+4.82\%} on WebWalkerQA compared to the  GRPO baseline.

A breakdown of the contributions reveals that DDPO provides a strong foundation, boosting accuracy by 2.69\% and 3.55\% on the two datasets, respectively, by mitigating reward conflicts. Furthermore, the integration of CANR yields an additional gain of +0.92\% on WebVLN and +1.27\% on WebWalkerQA. These incremental improvements are critical, as they confirm that the confidence-guided reflection mechanism effectively calibrates uncertain decisions at individual steps, preventing error propagation in long-horizon navigation tasks.

As visualized in Figure~\ref{fig:example}, StepGuard effectively mitigates intermediate errors by revising its decision through reflection, ensuring the final navigation success.

\subsection{Ablation Study}
\noindent\textbf{Impact of Reflection Strategy.}
To verify whether the performance gain stems from the reflection mechanism itself or the specific confidence-based triggering, we compare random triggering, always-on static reflection, and our adaptive strategy. As shown in Table \ref{tab:adaptive_vs_random}, the \textit{Random Triggering} strategy yields inferior performance compared to our \textit{Adaptive} approach (39.02\% vs. 39.83\% SR). Although the always-on strategy can occasionally improve over random triggering, it substantially increases the Trajectory Length (TL) to 7.33 and inflates the average inference time to 13.2 seconds. In contrast, our confidence-guided mechanism achieves higher success rates with a shorter TL (4.58) and lower runtime (8.2 seconds). This confirms that reflection is computationally expensive and potentially distracting if misused; it provides the most benefit only when applied selectively to steps where the agent exhibits high uncertainty.

\begin{table}[t]
\caption{Comparison of reflection strategies on WebVLN. Our adaptive approach achieves the best overall performance while avoiding the excessive trajectory length and runtime overhead introduced by always-on reflection. Time denotes the average inference time per trajectory (seconds).}
\vspace{-0.3cm}
\label{tab:adaptive_vs_random}
\centering
\small
\setlength{\tabcolsep}{2pt}
\begin{tabular*}{\columnwidth}{@{\extracolsep{\fill}}lccccccc}
\toprule
\multirow{2}{*}{\textbf{Strategy}} & \multirow{2}{*}{\textbf{SR}} & \multirow{2}{*}{\textbf{OSR}} & \multirow{2}{*}{\textbf{SPL}} & \multirow{2}{*}{\textbf{TL}} & \multicolumn{2}{c}{\textbf{WUPS}} & \multirow{2}{*}{\textbf{Time}} \\
\cmidrule(lr){6-7}
 &  &  &  &  & \textbf{@.9} & \textbf{@.0} & \\
\midrule
Random & 39.02 & 39.10 & 38.46 & 4.72 & 30.41 & 33.02 & 8.5 \\
Always-on & 39.37 & 39.52 & 37.84 & 7.33 & 30.89 & 34.40 & 13.2 \\
\textbf{CANR} & \textbf{39.83} & \textbf{39.94} & \textbf{39.21} & \textbf{4.58} & \textbf{31.28} & \textbf{33.82} & \textbf{8.2} \\
\bottomrule
\end{tabular*}

\end{table}

\begin{table}[t]
\caption{Confidence calibration analysis on WebVLN test set. The values represent the mean action probability averaged over all steps for successful versus failed episodes. The ``Gap'' denotes the margin between confidence in correct and incorrect predictions, serving as a metric for the model's self-assessment capability.}
\vspace{-0.3cm}

\label{tab:confidence_analysis}
\centering
\small

\setlength{\tabcolsep}{0pt} 

\begin{tabular*}{\columnwidth}{@{\extracolsep{\fill}}lccc}
\toprule
\multirow{2}{*}{\textbf{Method}} & \multicolumn{2}{c}{\textbf{Avg. Confidence}} & \multirow{2}{*}{\textbf{Gap ($\Delta$)}} \\
\cmidrule(lr){2-3}
 & \textbf{Correct Cases} & \textbf{Error Cases} & \\
\midrule
w/o CANR & 0.71 & 0.65 & 0.06 \\
\textbf{w/ CANR} & \textbf{0.74} & \textbf{0.57} & \textbf{0.17} \\
\bottomrule
\end{tabular*}

\vspace{-0.5cm}
\end{table}

\noindent\textbf{Confidence Calibration Analysis.} 
A core assumption underlying our CANR module is that the navigation confidence metric can distinguish reliable from unreliable decisions. To examine this, we analyze the distributions of confidence scores for correct and erroneous actions on the test set, comparing the standard model (w/o CANR training) with StepGuard (w/ CANR training).
As shown in Table \ref{tab:confidence_analysis}, CANR leads to substantially improved calibration. The baseline is overconfident, assigning high confidence even to erroneous actions (0.65), which yields a small discriminative gap ($\Delta=0.06$). In contrast, StepGuard reduces confidence for erroneous actions to 0.57 while preserving high confidence for correct actions (0.74). The enlarged gap ($\Delta=0.17$) suggests that the model learns to align its internal scalar confidence with the actual correctness of its policy, supporting the use of this metric as a trigger for adaptive reflection. Additional comparisons with entropy-based uncertainty estimation are provided in 
\hyperref[supp:sec.A]{Appendix A}.

\begin{table}[t]
\caption{Ablation study of reward components on \textbf{WebVLN}. "Inv." denotes the Invalid Action Rate (\%), indicating the percentage of steps where the model fails to generate executable actions.}
\vspace{-0.3cm}

\label{tab:ablation_webvln}
\centering
\small
\setlength{\tabcolsep}{0pt}
\begin{tabular*}{\columnwidth}{@{\extracolsep{\fill}}lcccc}
\toprule
\textbf{Configuration} & \textbf{SR} $\uparrow$ & \textbf{SPL} $\uparrow$ & \textbf{TL} $\downarrow$ & \textbf{Inv.} $\downarrow$ \\
\midrule
\textbf{Full Reward} & \textbf{39.83} & \textbf{39.21} & 4.58 & \textbf{0.45} \\
\hspace{1em} w/o $R_{rs}$ & 39.07 & 36.45 & 4.42 & 0.92 \\
\hspace{1em} w/o $R_{f}$ & 33.40 & 30.18 & 5.89 & 14.25 \\
\bottomrule
\end{tabular*}
\vspace{-0.3cm}

\end{table}

\begin{table}[t]
\caption{Ablation study of reward components on \textbf{WebWalkerQA}. We report the Overall Success Rate (SR), Action Count (A.C.), and Invalid Action Rate (Inv.).}
\vspace{-0.3cm}

\label{tab:ablation_webwalker}
\centering
\small
\setlength{\tabcolsep}{0pt}
\begin{tabular*}{\columnwidth}{@{\extracolsep{\fill}}lccc}
\toprule
\textbf{Configuration} & \textbf{SR} $\uparrow$ & \textbf{A.C.} $\downarrow$ & \textbf{Inv.} $\downarrow$ \\
\midrule
\textbf{Full Reward} & \textbf{42.41} & \textbf{4.68} & \textbf{0.62} \\
\hspace{1em} w/o $R_{rs}$ & 40.91 & 4.99 & 1.15 \\
\hspace{1em} w/o $R_{f}$ & 34.11 & 6.15 & 15.80 \\
\bottomrule
\end{tabular*}

\vspace{-0.65cm}
\end{table}

\noindent\textbf{Impact of Reward Components.} 
To assess the contribution of each reward component, we conduct ablation studies on WebVLN and WebWalkerQA, as detailed in Table \ref{tab:ablation_webvln} and Table \ref{tab:ablation_webwalker}.
The results show that the Format Reward ($R_f$) is fundamental to system stability, especially given the lightweight Qwen2.5-3B backbone. Removing $R_f$ leads to a sharp increase in the invalid action rate (peaking at 15.80\% on WebWalkerQA), substantially degrading overall success.
This confirms that explicit syntactic penalties are crucial for enforcing instruction adherence in smaller language models. Moreover, the Reasoning Step Reward ($R_{rs}$) plays a key role in strengthening the agent’s reflective ability by providing dense supervision over intermediate reasoning steps. The consistent decline in Success Rate across both datasets after removing $R_{rs}$ demonstrates that this reward signal is essential for aligning the agent’s internal reasoning process with effective decision-making and ensuring that self-correction mechanisms are actively used to achieve the navigation goal.


%% file: conclusion.tex
\section{
Conclusion}

In this paper, we present StepGuard, a new framework for robust web navigation that remedies single-step fragility through reward decoupling and adaptive calibration. By introducing Dynamic Dual-Policy Optimization (DDPO) and Confidence-Guided Adaptive Navigation Reflection (CANR), our method effectively mitigates reward conflict and error propagation during navigation. Extensive experiments on standard benchmarks demonstrate that StepGuard consistently improves both navigation and answer accuracy, establishing new state-of-the-art results and offering a principled direction for reliable web navigation.

%% file: limit.tex
\section{Limitations}

Dependence on Explicit Reward Shaping.
While our method effectively activates the navigation capabilities of the lightweight Qwen2.5-3B model, it relies heavily on dense reward engineering, particularly the format constraint ($R_f$). As observed in our studies, the model's intrinsic instruction-following ability is relatively weak compared to larger-scale foundation models (e.g., 70B+ parameters). Consequently, adapting our framework to tasks without well-defined validity signals or where dense supervision is unavailable may present convergence challenges.

%% file: appendix.tex
\appendix

\twocolumn[
  \phantomsection 
  \label{sec:appendix_start} 
  \begin{@twocolumnfalse} 
    \centering
    
    {
      \Large \textbf{
    \raisebox{-0.15cm}{\includegraphics[height=1.3em]{StepGuard_logo.png}}~ StepGuard: Guarding Web Navigation via Single-Step Calibration
      } \par
    }

    \vspace{0.5em}
    
    {\large \textit{
    Supplementary Materials} \par
    }
    \vspace{1em}
    
    {
      \small 
      \startcontents[appendices]
      \setcounter{tocdepth}{2} 
      \printcontents[appendices]{}{1}{\section*{Contents}}
    }
\vspace{1em}
\hrule height 1pt \vspace{2pt} \hrule height 0.4pt
\vspace{2em}

\end{@twocolumnfalse}
]

\section{Experiment Setup}
\label{supp:sec.A}
\subsection{Implementation Details}

\noindent\textbf{Training Details.} 
All experiments are implemented in PyTorch and conducted on a cluster of 10 NVIDIA GeForce RTX 3090 GPUs (24GB). 
Given the memory constraints of consumer-grade hardware, we employ full parameter fine-tuning rather than parameter-efficient methods (e.g., LoRA). To achieve this, we leverage DeepSpeed ZeRO Stage 3 \cite{rajbhandari2020zero} to partition optimizer states, gradients, and model parameters across GPUs, combined with Flash Attention 2 \cite{dao2023flashattention} and gradient checkpointing to further optimize memory utilization.
For the \textbf{WebVLN} task, the Qwen2.5-VL-3B-Instruct model is trained using the VLM-R1 framework. We employ Group Relative Policy Optimization (GRPO) with a group size of $G=8$ and a KL-divergence coefficient of $\beta=0.04$. The visual encoder processes images at their native resolution with a maximum token limit of 4,096. 
For the \textbf{WebWalkerQA} task, the Qwen2.5-VL-3B-Instruct model is fine-tuned using the Ms-Swift framework \cite{zhao2024swift}. Consistent with our visual alignment strategy, we apply GRPO with a group size of $G=8$ and a KL-divergence coefficient of $\beta=0.04$ to ensure stable policy updates against the reference model. To accommodate the verbose HTML observations and long-horizon history, we configure a maximum sequence length of 6,144 tokens, enabling the agent to process complex DOM tree structures without significant information loss.
Both models are optimized using AdamW \cite{loshchilov2017decoupled} with $\beta_1=0.9$, $\beta_2=0.95$, and a weight decay of 0.1. 
We utilize a cosine learning rate scheduler with a warmup ratio of 0.03 and a maximum learning rate of $2 \times 10^{-6}$ to ensure training stability. 
The global batch size is set to 64 via gradient accumulation, and all training is performed in BFloat16 precision.

\noindent\textbf{Datasets}
We evaluate our method on two benchmark datasets: (1) WebVLN, a web navigation dataset designed to evaluate the ability of agents to follow instructions and navigate through web pages to retrieve information. WebVLN contains a variety of queries that require agents to traverse different web pages systematically to answer questions based on the content of those pages. (2) WebWalkerQA, a multi-agent web navigation framework focused on simulating human-like web exploration. It contains 680 question-answer pairs derived from four real-world domains: education, conferences, organizations, and games, across over 1373 webpages. For our experiments, we focus on the single-source QA subset of WebWalkerQA for both training and evaluation.

\noindent\textbf{Metrics.} 
For WebVLN, we use the following evaluation metrics: Success Rate (SR), which measures the percentage of tasks where the agent successfully navigates to the target webpage; Oracle Success Rate (OSR), which measures how often the agent's performance matches the optimal trajectory; Success-weighted Path Length (SPL), which combines success rate and path efficiency; and Trajectory Length (TL), which measures the total number of steps taken by the agent. For question-answering (QA), we adopt Wu-Palmer Similarity (WUPS), which quantifies the semantic similarity between the predicted and ground-truth answers, using thresholds of 0.9 and 0.0.

For WebWalkerQA, we evaluate performance using question-answering accuracy (acc.), which measures the correctness of the model's responses, and Action Count (A.C.), which reflects the efficiency of the navigation process. Due to the varying lengths of the generated answers, we utilize GPT-4 as an evaluator to determine the correctness of responses by comparing them with the ground truth using the Chain-of-Thought (CoT) prompting strategy (Wei et al., 2022).

\noindent\textbf{Settings.}
For WebVLN, we follow the standard split protocol and partition the dataset into 60\% training, 10\% validation, and 30\% testing samples (i.e., 8,960/1,262/4,603). Since WebVLN involves visual webpage observations, we adopt Qwen2.5-VL-3B-Instruct as the backbone model. For WebWalkerQA, we focus on the single-source QA subset for training and evaluation, and adopt Qwen2.5-3B-Instruct as the backbone model given the absence of image modality in this dataset. All models are optimized with GRPO, and the reported numbers are averaged over three random seeds.

\subsection{Prompt Formulation and Reflective Instructions}
\label{sec:prompt_design}

To ensure the agent follows the reasoning requirements, we design a straightforward prompt template for standard navigation. Additionally, we introduce a set of concise reflection prompts for the CANR module to trigger immediate self-correction.

\noindent\textbf{Standard Navigation Prompt.} 
We construct the system instruction based on a fixed template that directly maps the environment state to the model input. As shown in Figure~\ref{fig:prompt_template}, the prompt consists of four essential parts:
\begin{itemize}
    \item \textbf{Role \& Objective:} Defines the agent as a shopping assistant and explicitly states the target product and user question.
    \item \textbf{Button List:} Provides the text description and unique IDs of all interactive buttons on the current page.
    \item \textbf{Constraint:} Restricts the agent to selecting a valid button ID or the \texttt{STOP} action.
    \item \textbf{Output Format:} Requires the agent to enclose the reasoning process in \texttt{<think>} tags and the final decision in \texttt{<answer>} tags. The answer must contain an \texttt{'action'} field (button ID or STOP) and an \texttt{'answer'} field (response text).
\end{itemize}

\begin{figure}[h]
\centering
\begin{tcolorbox}[colback=gray!10!white,colframe=black!75!black,title=Prompt Template]
\small
\textbf{Role:} You are a reasoning agent assistant for a website interface. \\
\textbf{Objective:} Navigate to the target product: "\texttt{\{target\}}" and answer the user's question: "\texttt{\{user\_request\}}". \\
\textbf{Observation:} Button list: \texttt{\{obs\_description\}} \\
\textbf{Constraint:} You can only select one button ID to jump to the next page. If the current page matches the target, stop and answer. \\
\textbf{Output Format:} Output the thinking process in \texttt{<think>...</think>} and answer in \texttt{<answer>...</answer>} tags in the following format: \texttt{<think>...</think><answer>['action': enum[button ID] or STOP, 'answer': 'no answer text' if action is a button ID else 'your answer to the user's question']</answer>}.
\end{tcolorbox}
\caption{The prompt template used for training and inference. It enforces a strict separation between reasoning traces and actionable outputs.}
\label{fig:prompt_template}
\end{figure}

\noindent\textbf{CANR Reflection Prompts.} 
When the CANR module triggers a reflection, we interrupt the standard generation process. To avoid complex instructions that might confuse the model, we use simple and direct prompts to force the agent to re-evaluate its current status. 
We design a pool of diverse reflection prompts to ensure generalization. At each trigger step, one prompt is randomly selected from the following list to append to the context:

\begin{itemize}
    \item \textit{"Wait. I need to closely re-examine the page details to ensure I have not overlooked any key attributes required by the target."} (Observation Check)
    \item \textit{"I must reflect on my action history and switch to a different search path to avoid potential loops or stagnation."} (Trajectory Check)
    \item \textit{"I should revisit the user's request to verify that my current progress strictly aligns with the specific goal."} (Goal Check)
    \item \textit{"I need to pause and evaluate whether a more optimal button exists to reach the target page faster."} (Efficiency Check)
\end{itemize}

These reflection prompts are strategically designed to address four distinct categories of cognitive failure often observed in web navigation agents:
(1) Observation Check targets the perception-reasoning gap, mitigating hallucinations where the agent overlooks fine-grained attributes in complex DOM structures;
(2) Trajectory Check addresses temporal stagnation, preventing the agent from falling into cyclic loops or local minima by forcing a review of historical actions;
(3) Goal Check counteracts instruction drift in long-horizon tasks, ensuring the agent remains aligned with the user's specific constraints rather than generic category matching;
(4) Efficiency Check discourages greedy behavior, prompting the agent to seek the optimal path rather than merely the first feasible action.

By randomly injecting these concise signals, we simulate a "wake-up call" that effectively breaks the agent's inertia and encourages it to generate a new, corrected chain of thought.

\begin{figure*}[t]
\centering
  \includegraphics[width=0.96\linewidth]{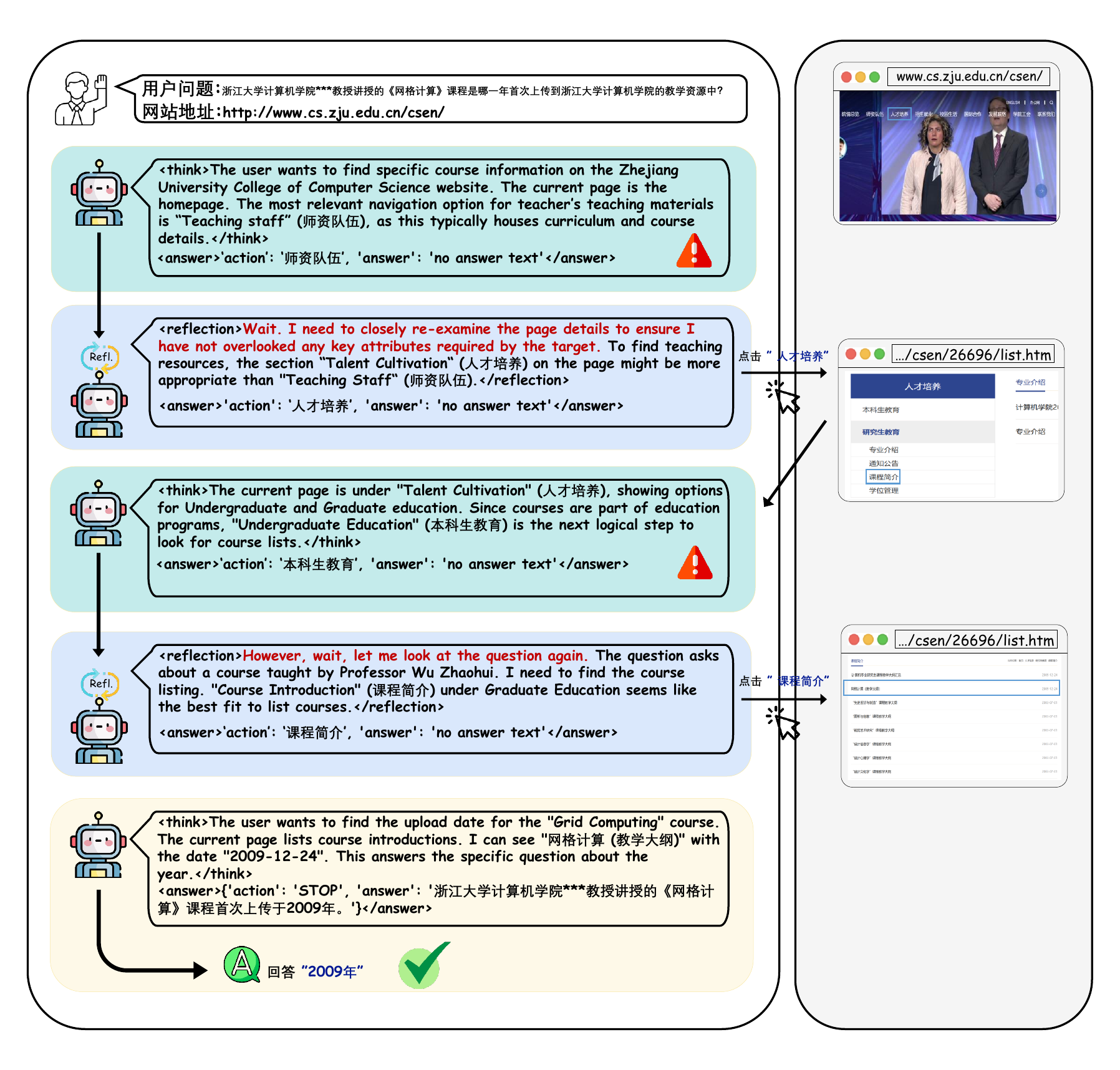}
  \vspace{-0.6cm}
\caption{Visualization of iterative self-correction on a challenging WebWalker sample. The figure demonstrates how StepGuard utilizes multiple reflection triggers, highlighted in red frames, to escape semantic traps and rectify its navigation path during a multi-step retrieval task.
\textit{English translation of the non-English text:} The user asks in which year the Grid Computing course taught by a professor at Zhejiang University College of Computer Science was first uploaded to the teaching resources platform. The relevant actions are ``Faculty'', ``Talent Cultivation'', ``Undergraduate Education'', and ``Course Introduction''. StepGuard finally answers the question correctly: 2009.}
\label{fig:appendix_case}
\end{figure*}

\section{Extended Analysis and Discussion}
\label{supp:sec.B}
\subsection{Sensitivity Analysis of Reflection Trigger}
\label{subsec:kappa_analysis}

We investigate the impact of the sensitivity coefficient $\kappa$ (in Eq.~\ref{eq:preflect}) on the agent's performance. As shown in Table~\ref{tab:ablation_kappa}, $\kappa$ acts as a critical threshold that balances reflection frequency and inference efficiency.

\begin{itemize}
    \item \textbf{Low $\kappa$ (1, 2):} The agent becomes \textit{over-sensitive}, triggering reflections even when relatively confident. This introduces stochastic noise into correct reasoning paths and increases computational overhead, resulting in lower SPL.
    \item \textbf{High $\kappa$ (6):} The mechanism becomes \textit{under-sensitive} (or inert). It fails to activate self-correction during low-confidence states, causing the performance to degrade towards the baseline.
    \item \textbf{Moderate $\kappa$ (4):} This setting achieves the optimal equilibrium. It creates a robust gating function that effectively filters out false positives while ensuring aggressive intervention when genuine uncertainty arises.
\end{itemize}

\begin{table}[h]
\centering
\small
\caption{Ablation study of the sensitivity coefficient $\kappa$ on the WebVLN dataset. The results demonstrate the robustness of our method: even with suboptimal $\kappa$ values, CANR consistently outperforms the strong \textbf{DDPO baseline}.}
\label{tab:ablation_kappa}
\setlength{\tabcolsep}{0pt} 
\begin{tabular*}{\columnwidth}{@{\extracolsep{\fill}}lcccc}
\toprule
\multirow{2}{*}{$\boldsymbol{\kappa}$ \textbf{Value}} & \multicolumn{2}{c}{\textbf{Validation}} & \multicolumn{2}{c}{\textbf{Test}} \\
\cmidrule(lr){2-3} \cmidrule(lr){4-5}
 & \textbf{SR} $\uparrow$ & \textbf{SPL} $\uparrow$ & \textbf{SR} $\uparrow$ & \textbf{SPL} $\uparrow$ \\
\midrule
\textit{Baseline (+DDPO)} & 41.64 & 41.05 & 38.58 & 37.99 \\
\midrule
1 \textit{(Over-sensitive)} & 42.05 & 41.38 & 39.17 & 38.45 \\
2 & 42.21 & 41.62 & 39.52 & 38.90 \\
\textbf{4 \textit{(Default)}} & \textbf{42.32} & \textbf{41.76} & \textbf{39.83} & \textbf{39.21} \\
6 \textit{(Under-sensitive)} & 41.89 & 41.14 & 38.92 & 38.13 \\
\bottomrule
\end{tabular*}
\end{table}

\subsection{Confidence Metric Validation}
\label{subsec:confidence_validation}

We first formalize the relationship between the confidence score in Eq.~\ref{eq:conf} and entropy. Let $\pi_t = \pi_\theta(\cdot \mid s_t)$ and let $U(C_t)$ denote the uniform distribution over the candidate action set $C_t$. Then
\begin{equation}
\begin{aligned}
D_{KL}(\pi_t\|U(C_t))
&= \sum_{a \in C_t} \pi_t(a)
   \log \frac{\pi_t(a)}{1/|C_t|} \\
&= \sum_{a \in C_t} \pi_t(a) \log \pi_t(a) \\
&\quad + \sum_{a \in C_t} \pi_t(a) \log |C_t| \\
&= -H(\pi_t) + \log |C_t|.
\end{aligned}
\end{equation}
Hence, for a fixed candidate set size, KL divergence and entropy differ only by a constant shift and therefore induce the same ranking over uncertain states. Motivated by this equivalence, we compare the KL-divergence score in Eq.~\ref{eq:conf} with normalized entropy on WebVLN. Table~\ref{tab:entropy_vs_kl} confirms this expectation: both metrics yield closely matched performance, while KL divergence provides slightly stronger overall results in our final setting.

\begin{table*}[t]
\centering
\small
\setlength{\tabcolsep}{4pt}
\begin{adjustbox}{max width=\textwidth}
\begin{tabular}{lcccccccccccc}
\toprule
\multirow{2}{*}{\textbf{Method}} & \multicolumn{6}{c}{\textbf{Val}} & \multicolumn{6}{c}{\textbf{Test}} \\
\cmidrule(lr){2-7} \cmidrule(lr){8-13}
 & \textbf{SR} & \textbf{OSR} & \textbf{SPL} & \textbf{TL} & \textbf{WUPS0.9} & \textbf{WUPS0.0} & \textbf{SR} & \textbf{OSR} & \textbf{SPL} & \textbf{TL} & \textbf{WUPS0.9} & \textbf{WUPS0.0} \\
\midrule
Normalized Entropy & 42.40 & 42.45 & 41.68 & 5.09 & 29.99 & 35.16 & 39.43 & 39.58 & 38.96 & 4.72 & 30.87 & 33.50 \\
KL divergence & 42.32 & 42.41 & 41.76 & 4.99 & 30.63 & 35.92 & 39.83 & 39.94 & 39.21 & 4.58 & 31.28 & 33.82 \\
\bottomrule
\end{tabular}
\end{adjustbox}
\caption{Comparison between normalized entropy and KL divergence as confidence metrics on WebVLN. The two formulations yield closely aligned outcomes, supporting the use of KL divergence as our default uncertainty signal.}
\label{tab:entropy_vs_kl}
\end{table*}

We also examine whether the confidence score tracks task difficulty on WebWalkerQA. Using the untrained Qwen2.5-3B model, we compute the average confidence on the Easy, Medium, and Hard subsets. As shown in Table~\ref{tab:difficulty_confidence}, confidence decreases monotonically with difficulty, indicating that the metric captures meaningful variation in decision uncertainty before any CANR optimization is applied.

\begin{table}[h]
\centering
\small
\setlength{\tabcolsep}{0pt}
\begin{tabular*}{\columnwidth}{@{\extracolsep{\fill}}ccc}
\toprule
\textbf{Easy} & \textbf{Medium} & \textbf{Hard} \\
\midrule
0.67 & 0.63 & 0.51 \\
\bottomrule
\end{tabular*}
\caption{Average confidence scores on the Easy, Medium, and Hard subsets of WebWalkerQA using the untrained Qwen2.5-3B model. Lower confidence is associated with more difficult tasks.}
\label{tab:difficulty_confidence}
\end{table}

\subsection{Runtime Analysis of Reflection Strategies}
\label{subsec:runtime_analysis}

We further quantify the computational overhead of different reflection strategies on WebVLN. Table~\ref{tab:runtime_reflection} reports the average inference time per trajectory, measured with Qwen2.5-VL-3B on a single RTX 3090 GPU. While always-on reflection can occasionally improve over random triggering, it incurs a substantial cost increase due to excessive reasoning steps. In contrast, CANR preserves the best overall task performance while keeping the runtime close to the random baseline.

\begin{table}[h]
\centering
\small
\setlength{\tabcolsep}{0pt}
\begin{tabular*}{\columnwidth}{@{\extracolsep{\fill}}lcc}
\toprule
\textbf{Strategy} & \textbf{TL} $\downarrow$ & \textbf{Avg. Time (s)} $\downarrow$ \\
\midrule
Random & 4.72 & 8.5 \\
Always-on (Static) & 7.33 & 13.2 \\
CANR (Adaptive) & \textbf{4.58} & \textbf{8.2} \\
\bottomrule
\end{tabular*}
\caption{Runtime comparison of reflection strategies on WebVLN. Adaptive CANR reduces both interaction steps and inference time relative to always-on reflection.}
\label{tab:runtime_reflection}
\end{table}

\subsection{Qualitative Analysis of Reasoning Trajectories}
\label{subsec:case_studies}

To illustrate the robustness of StepGuard in handling complex, long-horizon tasks, we present a challenging qualitative example from the WebWalker dataset in Figure~\ref{fig:appendix_case}. This case involves deep information retrieval where the target information is nested within ambiguous sub-menus, often leading standard agents into semantic traps.

As shown in the visualization, the agent initially exhibits a tendency to rely on superficial semantic associations, planning to click ``Teaching Staff'' based on the professor's name. However, the CANR module detects a potential misalignment and triggers a reflection. This ``wake-up call'' forces the agent to re-evaluate the page content against the specific goal (finding course materials), leading it to correctly choose ``Talent Cultivation''.

Subsequently, when facing another ambiguous choice between ``Undergraduate Education'' and ``Course Introduction'', a second reflection cycle prevents a suboptimal click. By revisiting the user's specific request for a course syllabus, the agent rectifies its trajectory and successfully navigates to the target page to retrieve the answer (``2009''). This episode demonstrates StepGuard's capability to perform iterative self-correction, effectively escaping Local optimal solutions where baseline models typically fail.

\section{Statements}
\label{supp:sec.C}
\subsection{Ethical Considerations}
 As autonomous web agents increasingly interact with dynamic environments, ensuring operational safety is critical to prevent unintended actions. Our proposed framework, StepGuard, mitigates these risks by incorporating a confidence-guided reflection mechanism (CANR) that detects uncertainty and corrects potential errors before execution, thereby enhancing system reliability. Furthermore, our approach contributes to sustainable AI by demonstrating that a lightweight 3B-parameter model can achieve performance comparable to large-scale foundation models (e.g., 72B), significantly reducing the computational energy required for training and deployment. Finally, our experiments utilize established public benchmarks (WebVLN and WebWalkerQA) that do not contain personally identifiable information.

\subsection{Reproducibility Statement}
To facilitate the reproducibility of our work, we provide comprehensive implementation details and experimental configurations. The complete source code will be made publicly available at github upon acceptance.

%% file: custom.bib
@article{qwenvl,
  author       = {Shuai Bai and
                  Keqin Chen and
                  Xuejing Liu and
                  Jialin Wang and
                  Wenbin Ge and
                  Sibo Song and
                  Kai Dang and
                  Peng Wang and
                  Shijie Wang and
                  Jun Tang and
                  Humen Zhong and
                  Yuanzhi Zhu and
                  Ming{-}Hsuan Yang and
                  Zhaohai Li and
                  Jianqiang Wan and
                  Pengfei Wang and
                  Wei Ding and
                  Zheren Fu and
                  Yiheng Xu and
                  Jiabo Ye and
                  Xi Zhang and
                  Tianbao Xie and
                  Zesen Cheng and
                  Hang Zhang and
                  Zhibo Yang and
                  Haiyang Xu and
                  Junyang Lin},
  title        = {Qwen2.5-VL Technical Report},
  journal      = {CoRR},
  volume       = {abs/2502.13923},
  year         = {2025},
  url          = {https://doi.org/10.48550/arXiv.2502.13923},
  doi          = {10.48550/ARXIV.2502.13923},
  eprinttype    = {arXiv},
  eprint       = {2502.13923},
  bibsource    = {dblp computer science bibliography, https://dblp.org}
}

@article{InternVL,
  author       = {Zhe Chen and
                  Jiannan Wu and
                  Wenhai Wang and
                  Weijie Su and
                  Guo Chen and
                  Sen Xing and
                  Muyan Zhong and
                  Qinglong Zhang and
                  Xizhou Zhu and
                  Lewei Lu and
                  Bin Li and
                  Ping Luo and
                  Tong Lu and
                  Yu Qiao and
                  Jifeng Dai},
  title        = {InternVL: Scaling up Vision Foundation Models and Aligning for Generic
                  Visual-Linguistic Tasks},
  journal      = {CoRR},
  volume       = {abs/2312.14238},
  year         = {2023},
  url          = {https://doi.org/10.48550/arXiv.2312.14238},
  doi          = {10.48550/ARXIV.2312.14238},
  eprinttype    = {arXiv},
  eprint       = {2312.14238},
  bibsource    = {dblp computer science bibliography, https://dblp.org}
}

@inproceedings{MiniWoB++,
  author       = {Evan Zheran Liu and
                  Kelvin Guu and
                  Panupong Pasupat and
                  Tianlin Shi and
                  Percy Liang},
  title        = {Reinforcement Learning on Web Interfaces using Workflow-Guided Exploration},
  booktitle    = {6th International Conference on Learning Representations, {ICLR} 2018,
                  Vancouver, BC, Canada, April 30 - May 3, 2018, Conference Track Proceedings},
  publisher    = {OpenReview.net},
  year         = {2018},
  url          = {https://openreview.net/forum?id=ryTp3f-0-},
  bibsource    = {dblp computer science bibliography, https://dblp.org}
}

@inproceedings{RUSS,
  author       = {Nancy Xu and
                  Sam Masling and
                  Michael Du and
                  Giovanni Campagna and
                  Larry Heck and
                  James A. Landay and
                  Monica Lam},
  editor       = {Kristina Toutanova and
                  Anna Rumshisky and
                  Luke Zettlemoyer and
                  Dilek Hakkani{-}T{\"{u}}r and
                  Iz Beltagy and
                  Steven Bethard and
                  Ryan Cotterell and
                  Tanmoy Chakraborty and
                  Yichao Zhou},
  title        = {Grounding Open-Domain Instructions to Automate Web Support Tasks},
  booktitle    = {Proceedings of the 2021 Conference of the North American Chapter of
                  the Association for Computational Linguistics: Human Language Technologies,
                  {NAACL-HLT} 2021, Online, June 6-11, 2021},
  pages        = {1022--1032},
  publisher    = {Association for Computational Linguistics},
  year         = {2021},
  url          = {https://doi.org/10.18653/v1/2021.naacl-main.80},
  doi          = {10.18653/V1/2021.NAACL-MAIN.80},
  bibsource    = {dblp computer science bibliography, https://dblp.org}
}

@inproceedings{FLIN,
  author       = {Sahisnu Mazumder and
                  Oriana Riva},
  editor       = {Kristina Toutanova and
                  Anna Rumshisky and
                  Luke Zettlemoyer and
                  Dilek Hakkani{-}T{\"{u}}r and
                  Iz Beltagy and
                  Steven Bethard and
                  Ryan Cotterell and
                  Tanmoy Chakraborty and
                  Yichao Zhou},
  title        = {{FLIN:} {A} Flexible Natural Language Interface for Web Navigation},
  booktitle    = {Proceedings of the 2021 Conference of the North American Chapter of
                  the Association for Computational Linguistics: Human Language Technologies,
                  {NAACL-HLT} 2021, Online, June 6-11, 2021},
  pages        = {2777--2788},
  publisher    = {Association for Computational Linguistics},
  year         = {2021},
  url          = {https://doi.org/10.18653/v1/2021.naacl-main.222},
  doi          = {10.18653/V1/2021.NAACL-MAIN.222},
  bibsource    = {dblp computer science bibliography, https://dblp.org}
}

@inproceedings{WebQA,
  author       = {Yingshan Chang and
                  Yonatan Bisk},
  editor       = {Douwe Kiela and
                  Marco Ciccone and
                  Barbara Caputo},
  title        = {WebQA: {A} Multimodal Multihop NeurIPS Challenge},
  booktitle    = {NeurIPS 2021 Competitions and Demonstrations Track, 6-14 December
                  2021, Online},
  series       = {Proceedings of Machine Learning Research},
  volume       = {176},
  pages        = {232--245},
  publisher    = {{PMLR}},
  year         = {2021},
  url          = {https://proceedings.mlr.press/v176/chang22a.html},
  bibsource    = {dblp computer science bibliography, https://dblp.org}
}

@inproceedings{ScreenQA,
  author       = {Yu{-}Chung Hsiao and
                  Fedir Zubach and
                  Gilles Baechler and
                  Srinivas Sunkara and
                  Victor Carbune and
                  Jason Lin and
                  Maria Wang and
                  Yun Zhu and
                  Jindong Chen},
  editor       = {Luis Chiruzzo and
                  Alan Ritter and
                  Lu Wang},
  title        = {ScreenQA: Large-Scale Question-Answer Pairs Over Mobile App Screenshots},
  booktitle    = {Proceedings of the 2025 Conference of the Nations of the Americas
                  Chapter of the Association for Computational Linguistics: Human Language
                  Technologies, {NAACL} 2025 - Volume 1: Long Papers, Albuquerque, New
                  Mexico, USA, April 29 - May 4, 2025},
  pages        = {9427--9452},
  publisher    = {Association for Computational Linguistics},
  year         = {2025},
  url          = {https://doi.org/10.18653/v1/2025.naacl-long.477},
  doi          = {10.18653/V1/2025.NAACL-LONG.477},
  bibsource    = {dblp computer science bibliography, https://dblp.org}
}

@inproceedings{mind2web,
  author       = {Xiang Deng and
                  Yu Gu and
                  Boyuan Zheng and
                  Shijie Chen and
                  Samual Stevens and
                  Boshi Wang and
                  Huan Sun and
                  Yu Su},
  editor       = {Alice Oh and
                  Tristan Naumann and
                  Amir Globerson and
                  Kate Saenko and
                  Moritz Hardt and
                  Sergey Levine},
  title        = {Mind2Web: Towards a Generalist Agent for the Web},
  booktitle    = {Advances in Neural Information Processing Systems 36: Annual Conference
                  on Neural Information Processing Systems 2023, NeurIPS 2023, New Orleans,
                  LA, USA, December 10 - 16, 2023},
  year         = {2023},
  url          = {http://papers.nips.cc/paper\_files/paper/2023/hash/5950bf290a1570ea401bf98882128160-Abstract-Datasets\_and\_Benchmarks.html},
  bibsource    = {dblp computer science bibliography, https://dblp.org}
}

@inproceedings{WebArena,
  author       = {Shuyan Zhou and
                  Frank F. Xu and
                  Hao Zhu and
                  Xuhui Zhou and
                  Robert Lo and
                  Abishek Sridhar and
                  Xianyi Cheng and
                  Tianyue Ou and
                  Yonatan Bisk and
                  Daniel Fried and
                  Uri Alon and
                  Graham Neubig},
  title        = {WebArena: {A} Realistic Web Environment for Building Autonomous Agents},
  booktitle    = {The Twelfth International Conference on Learning Representations,
                  {ICLR} 2024, Vienna, Austria, May 7-11, 2024},
  publisher    = {OpenReview.net},
  year         = {2024},
  url          = {https://openreview.net/forum?id=oKn9c6ytLx},
  bibsource    = {dblp computer science bibliography, https://dblp.org}
}

@inproceedings{WebShop,
  author       = {Shunyu Yao and
                  Howard Chen and
                  John Yang and
                  Karthik Narasimhan},
  editor       = {Sanmi Koyejo and
                  S. Mohamed and
                  A. Agarwal and
                  Danielle Belgrave and
                  K. Cho and
                  A. Oh},
  title        = {WebShop: Towards Scalable Real-World Web Interaction with Grounded
                  Language Agents},
  booktitle    = {Advances in Neural Information Processing Systems 35: Annual Conference
                  on Neural Information Processing Systems 2022, NeurIPS 2022, New Orleans,
                  LA, USA, November 28 - December 9, 2022},
  year         = {2022},
  url          = {http://papers.nips.cc/paper\_files/paper/2022/hash/82ad13ec01f9fe44c01cb91814fd7b8c-Abstract-Conference.html},
  bibsource    = {dblp computer science bibliography, https://dblp.org}
}

@inproceedings{SeeAct,
  author       = {Boyuan Zheng and
                  Boyu Gou and
                  Jihyung Kil and
                  Huan Sun and
                  Yu Su},
  title        = {GPT-4V(ision) is a Generalist Web Agent, if Grounded},
  booktitle    = {Forty-first International Conference on Machine Learning, {ICML} 2024,
                  Vienna, Austria, July 21-27, 2024},
  publisher    = {OpenReview.net},
  year         = {2024},
  url          = {https://openreview.net/forum?id=piecKJ2DlB},
  bibsource    = {dblp computer science bibliography, https://dblp.org}
}

@inproceedings{Synapse,
  author       = {Longtao Zheng and
                  Rundong Wang and
                  Xinrun Wang and
                  Bo An},
  title        = {Synapse: Trajectory-as-Exemplar Prompting with Memory for Computer
                  Control},
  booktitle    = {The Twelfth International Conference on Learning Representations,
                  {ICLR} 2024, Vienna, Austria, May 7-11, 2024},
  publisher    = {OpenReview.net},
  year         = {2024},
  url          = {https://openreview.net/forum?id=Pc8AU1aF5e},
  bibsource    = {dblp computer science bibliography, https://dblp.org}
}

@inproceedings{DigiRL,
  author       = {Hao Bai and
                  Yifei Zhou and
                  Jiayi Pan and
                  Mert Cemri and
                  Alane Suhr and
                  Sergey Levine and
                  Aviral Kumar},
  editor       = {Amir Globersons and
                  Lester Mackey and
                  Danielle Belgrave and
                  Angela Fan and
                  Ulrich Paquet and
                  Jakub M. Tomczak and
                  Cheng Zhang},
  title        = {DigiRL: Training In-The-Wild Device-Control Agents with Autonomous
                  Reinforcement Learning},
  booktitle    = {Advances in Neural Information Processing Systems 38: Annual Conference
                  on Neural Information Processing Systems 2024, NeurIPS 2024, Vancouver,
                  BC, Canada, December 10 - 15, 2024},
  year         = {2024},
  url          = {http://papers.nips.cc/paper\_files/paper/2024/hash/1704ddd0bb89f159dfe609b32c889995-Abstract-Conference.html},
  bibsource    = {dblp computer science bibliography, https://dblp.org}
}

@inproceedings{Agent-FLAN,
  author       = {Zehui Chen and
                  Kuikun Liu and
                  Qiuchen Wang and
                  Wenwei Zhang and
                  Jiangning Liu and
                  Dahua Lin and
                  Kai Chen and
                  Feng Zhao},
  editor       = {Lun{-}Wei Ku and
                  Andre Martins and
                  Vivek Srikumar},
  title        = {Agent-FLAN: Designing Data and Methods of Effective Agent Tuning for
                  Large Language Models},
  booktitle    = {Findings of the Association for Computational Linguistics, {ACL} 2024,
                  Bangkok, Thailand and virtual meeting, August 11-16, 2024},
  pages        = {9354--9366},
  publisher    = {Association for Computational Linguistics},
  year         = {2024},
  url          = {https://doi.org/10.18653/v1/2024.findings-acl.557},
  doi          = {10.18653/V1/2024.FINDINGS-ACL.557},
  bibsource    = {dblp computer science bibliography, https://dblp.org}
}

@inproceedings{WebRL,
  author       = {Zehan Qi and
                  Xiao Liu and
                  Iat Long Iong and
                  Hanyu Lai and
                  Xueqiao Sun and
                  Jiadai Sun and
                  Xinyue Yang and
                  Yu Yang and
                  Shuntian Yao and
                  Wei Xu and
                  Jie Tang and
                  Yuxiao Dong},
  title        = {WebRL: Training {LLM} Web Agents via Self-Evolving Online Curriculum
                  Reinforcement Learning},
  booktitle    = {The Thirteenth International Conference on Learning Representations,
                  {ICLR} 2025, Singapore, April 24-28, 2025},
  publisher    = {OpenReview.net},
  year         = {2025},
  url          = {https://openreview.net/forum?id=oVKEAFjEqv},
  bibsource    = {dblp computer science bibliography, https://dblp.org}
}

@inproceedings{AutoWebGLM,
  author       = {Hanyu Lai and
                  Xiao Liu and
                  Iat Long Iong and
                  Shuntian Yao and
                  Yuxuan Chen and
                  Pengbo Shen and
                  Hao Yu and
                  Hanchen Zhang and
                  Xiaohan Zhang and
                  Yuxiao Dong and
                  Jie Tang},
  editor       = {Ricardo Baeza{-}Yates and
                  Francesco Bonchi},
  title        = {AutoWebGLM: {A} Large Language Model-based Web Navigating Agent},
  booktitle    = {Proceedings of the 30th {ACM} {SIGKDD} Conference on Knowledge Discovery
                  and Data Mining, {KDD} 2024, Barcelona, Spain, August 25-29, 2024},
  pages        = {5295--5306},
  publisher    = {{ACM}},
  year         = {2024},
  url          = {https://doi.org/10.1145/3637528.3671620},
  doi          = {10.1145/3637528.3671620},
  bibsource    = {dblp computer science bibliography, https://dblp.org}
}

@article{WebGPT,
  author       = {Reiichiro Nakano and
                  Jacob Hilton and
                  Suchir Balaji and
                  Jeff Wu and
                  Long Ouyang and
                  Christina Kim and
                  Christopher Hesse and
                  Shantanu Jain and
                  Vineet Kosaraju and
                  William Saunders and
                  Xu Jiang and
                  Karl Cobbe and
                  Tyna Eloundou and
                  Gretchen Krueger and
                  Kevin Button and
                  Matthew Knight and
                  Benjamin Chess and
                  John Schulman},
  title        = {WebGPT: Browser-assisted question-answering with human feedback},
  journal      = {CoRR},
  volume       = {abs/2112.09332},
  year         = {2021},
  url          = {https://arxiv.org/abs/2112.09332},
  eprinttype    = {arXiv},
  eprint       = {2112.09332},
  bibsource    = {dblp computer science bibliography, https://dblp.org}
}

@article{CPAS,
  author       = {Inioluwa Deborah Raji and
                  Roel Dobbe},
  title        = {Concrete Problems in {AI} Safety, Revisited},
  journal      = {CoRR},
  volume       = {abs/2401.10899},
  year         = {2024},
  url          = {https://doi.org/10.48550/arXiv.2401.10899},
  doi          = {10.48550/ARXIV.2401.10899},
  eprinttype    = {arXiv},
  eprint       = {2401.10899},
  bibsource    = {dblp computer science bibliography, https://dblp.org}
}

@inproceedings{ARAG,
  author       = {Zhengbao Jiang and
                  Frank F. Xu and
                  Luyu Gao and
                  Zhiqing Sun and
                  Qian Liu and
                  Jane Dwivedi{-}Yu and
                  Yiming Yang and
                  Jamie Callan and
                  Graham Neubig},
  editor       = {Houda Bouamor and
                  Juan Pino and
                  Kalika Bali},
  title        = {Active Retrieval Augmented Generation},
  booktitle    = {Proceedings of the 2023 Conference on Empirical Methods in Natural
                  Language Processing, {EMNLP} 2023, Singapore, December 6-10, 2023},
  pages        = {7969--7992},
  publisher    = {Association for Computational Linguistics},
  year         = {2023},
  url          = {https://doi.org/10.18653/v1/2023.emnlp-main.495},
  doi          = {10.18653/V1/2023.EMNLP-MAIN.495},
  bibsource    = {dblp computer science bibliography, https://dblp.org}
}

@article{Multi-Objective-RL-Guide,
  author       = {Conor F. Hayes and
                  Roxana Radulescu and
                  Eugenio Bargiacchi and
                  Johan K{\"{a}}llstr{\"{o}}m and
                  Matthew Macfarlane and
                  Mathieu Reymond and
                  Timothy Verstraeten and
                  Luisa M. Zintgraf and
                  Richard Dazeley and
                  Fredrik Heintz and
                  Enda Howley and
                  Athirai A. Irissappane and
                  Patrick Mannion and
                  Ann Now{\'{e}} and
                  Gabriel de Oliveira Ramos and
                  Marcello Restelli and
                  Peter Vamplew and
                  Diederik M. Roijers},
  title        = {A practical guide to multi-objective reinforcement learning and planning},
  journal      = {Auton. Agents Multi Agent Syst.},
  volume       = {36},
  number       = {1},
  pages        = {26},
  year         = {2022},
  url          = {https://doi.org/10.1007/s10458-022-09552-y},
  doi          = {10.1007/S10458-022-09552-Y},
  bibsource    = {dblp computer science bibliography, https://dblp.org}
}

@inproceedings{Reflexion,
  author       = {Noah Shinn and
                  Federico Cassano and
                  Ashwin Gopinath and
                  Karthik Narasimhan and
                  Shunyu Yao},
  editor       = {Alice Oh and
                  Tristan Naumann and
                  Amir Globerson and
                  Kate Saenko and
                  Moritz Hardt and
                  Sergey Levine},
  title        = {Reflexion: language agents with verbal reinforcement learning},
  booktitle    = {Advances in Neural Information Processing Systems 36: Annual Conference
                  on Neural Information Processing Systems 2023, NeurIPS 2023, New Orleans,
                  LA, USA, December 10 - 16, 2023},
  year         = {2023},
  url          = {http://papers.nips.cc/paper\_files/paper/2023/hash/1b44b878bb782e6954cd888628510e90-Abstract-Conference.html},
  bibsource    = {dblp computer science bibliography, https://dblp.org}
}

@inproceedings{LATS,
  author       = {Andy Zhou and
                  Kai Yan and
                  Michal Shlapentokh{-}Rothman and
                  Haohan Wang and
                  Yu{-}Xiong Wang},
  title        = {Language Agent Tree Search Unifies Reasoning, Acting, and Planning
                  in Language Models},
  booktitle    = {Forty-first International Conference on Machine Learning, {ICML} 2024,
                  Vienna, Austria, July 21-27, 2024},
  publisher    = {OpenReview.net},
  year         = {2024},
  url          = {https://openreview.net/forum?id=njwv9BsGHF},
  bibsource    = {dblp computer science bibliography, https://dblp.org}
}

@inproceedings{PlanBench,
  author       = {Karthik Valmeekam and
                  Matthew Marquez and
                  Alberto Olmo Hernandez and
                  Sarath Sreedharan and
                  Subbarao Kambhampati},
  editor       = {Alice Oh and
                  Tristan Naumann and
                  Amir Globerson and
                  Kate Saenko and
                  Moritz Hardt and
                  Sergey Levine},
  title        = {PlanBench: An Extensible Benchmark for Evaluating Large Language Models
                  on Planning and Reasoning about Change},
  booktitle    = {Advances in Neural Information Processing Systems 36: Annual Conference
                  on Neural Information Processing Systems 2023, NeurIPS 2023, New Orleans,
                  LA, USA, December 10 - 16, 2023},
  year         = {2023},
  url          = {http://papers.nips.cc/paper\_files/paper/2023/hash/7a92bcdede88c7afd108072faf5485c8-Abstract-Datasets\_and\_Benchmarks.html},
  bibsource    = {dblp computer science bibliography, https://dblp.org}
}

@inproceedings{Just-Ask-for-Calibration,
  author       = {Katherine Tian and
                  Eric Mitchell and
                  Allan Zhou and
                  Archit Sharma and
                  Rafael Rafailov and
                  Huaxiu Yao and
                  Chelsea Finn and
                  Christopher D. Manning},
  editor       = {Houda Bouamor and
                  Juan Pino and
                  Kalika Bali},
  title        = {Just Ask for Calibration: Strategies for Eliciting Calibrated Confidence
                  Scores from Language Models Fine-Tuned with Human Feedback},
  booktitle    = {Proceedings of the 2023 Conference on Empirical Methods in Natural
                  Language Processing, {EMNLP} 2023, Singapore, December 6-10, 2023},
  pages        = {5433--5442},
  publisher    = {Association for Computational Linguistics},
  year         = {2023},
  url          = {https://doi.org/10.18653/v1/2023.emnlp-main.330},
  doi          = {10.18653/V1/2023.EMNLP-MAIN.330},
  bibsource    = {dblp computer science bibliography, https://dblp.org}
}

@inproceedings{Self-Consistency,
  author       = {Xuezhi Wang and
                  Jason Wei and
                  Dale Schuurmans and
                  Quoc V. Le and
                  Ed H. Chi and
                  Sharan Narang and
                  Aakanksha Chowdhery and
                  Denny Zhou},
  title        = {Self-Consistency Improves Chain of Thought Reasoning in Language Models},
  booktitle    = {The Eleventh International Conference on Learning Representations,
                  {ICLR} 2023, Kigali, Rwanda, May 1-5, 2023},
  publisher    = {OpenReview.net},
  year         = {2023},
  url          = {https://openreview.net/forum?id=1PL1NIMMrw},
  bibsource    = {dblp computer science bibliography, https://dblp.org}
}

@inproceedings{AgentBench,
  author       = {Xiao Liu and
                  Hao Yu and
                  Hanchen Zhang and
                  Yifan Xu and
                  Xuanyu Lei and
                  Hanyu Lai and
                  Yu Gu and
                  Hangliang Ding and
                  Kaiwen Men and
                  Kejuan Yang and
                  Shudan Zhang and
                  Xiang Deng and
                  Aohan Zeng and
                  Zhengxiao Du and
                  Chenhui Zhang and
                  Sheng Shen and
                  Tianjun Zhang and
                  Yu Su and
                  Huan Sun and
                  Minlie Huang and
                  Yuxiao Dong and
                  Jie Tang},
  title        = {AgentBench: Evaluating LLMs as Agents},
  booktitle    = {The Twelfth International Conference on Learning Representations,
                  {ICLR} 2024, Vienna, Austria, May 7-11, 2024},
  publisher    = {OpenReview.net},
  year         = {2024},
  url          = {https://openreview.net/forum?id=zAdUB0aCTQ},
  bibsource    = {dblp computer science bibliography, https://dblp.org}
}

@inproceedings{VisualWebArena,
  author       = {Jing Yu Koh and
                  Robert Lo and
                  Lawrence Jang and
                  Vikram Duvvur and
                  Ming Chong Lim and
                  Po{-}Yu Huang and
                  Graham Neubig and
                  Shuyan Zhou and
                  Russ Salakhutdinov and
                  Daniel Fried},
  editor       = {Lun{-}Wei Ku and
                  Andre Martins and
                  Vivek Srikumar},
  title        = {VisualWebArena: Evaluating Multimodal Agents on Realistic Visual Web
                  Tasks},
  booktitle    = {Proceedings of the 62nd Annual Meeting of the Association for Computational
                  Linguistics (Volume 1: Long Papers), {ACL} 2024, Bangkok, Thailand,
                  August 11-16, 2024},
  pages        = {881--905},
  publisher    = {Association for Computational Linguistics},
  year         = {2024},
  url          = {https://doi.org/10.18653/v1/2024.acl-long.50},
  doi          = {10.18653/V1/2024.ACL-LONG.50},
  bibsource    = {dblp computer science bibliography, https://dblp.org}
}

@article{GPT4TR,
  author    = {OpenAI},
  title     = {GPT-4 Technical Report},
  journal   = {CoRR},
  year      = {2023},
url={https://arxiv.org/abs/2303.08774},
}

@article{CoALA,
  author       = {Theodore R. Sumers and
                  Shunyu Yao and
                  Karthik Narasimhan and
                  Thomas L. Griffiths},
  title        = {Cognitive Architectures for Language Agents},
  journal      = {Trans. Mach. Learn. Res.},
  volume       = {2024},
  year         = {2024},
  url          = {https://openreview.net/forum?id=1i6ZCvflQJ},
  bibsource    = {dblp computer science bibliography, https://dblp.org}
}

@inproceedings{ExpeL,
  author       = {Andrew Zhao and
                  Daniel Huang and
                  Quentin Xu and
                  Matthieu Lin and
                  Yong{-}Jin Liu and
                  Gao Huang},
  editor       = {Michael J. Wooldridge and
                  Jennifer G. Dy and
                  Sriraam Natarajan},
  title        = {ExpeL: {LLM} Agents Are Experiential Learners},
  booktitle    = {Thirty-Eighth {AAAI} Conference on Artificial Intelligence, {AAAI}
                  2024, Thirty-Sixth Conference on Innovative Applications of Artificial
                  Intelligence, {IAAI} 2024, Fourteenth Symposium on Educational Advances
                  in Artificial Intelligence, {EAAI} 2014, February 20-27, 2024, Vancouver,
                  Canada},
  pages        = {19632--19642},
  publisher    = {{AAAI} Press},
  year         = {2024},
  url          = {https://doi.org/10.1609/aaai.v38i17.29936},
  doi          = {10.1609/AAAI.V38I17.29936},
  bibsource    = {dblp computer science bibliography, https://dblp.org}
}

@inproceedings{WebVoyager,
  author       = {Hongliang He and
                  Wenlin Yao and
                  Kaixin Ma and
                  Wenhao Yu and
                  Yong Dai and
                  Hongming Zhang and
                  Zhenzhong Lan and
                  Dong Yu},
  editor       = {Lun{-}Wei Ku and
                  Andre Martins and
                  Vivek Srikumar},
  title        = {WebVoyager: Building an End-to-End Web Agent with Large Multimodal
                  Models},
  booktitle    = {Proceedings of the 62nd Annual Meeting of the Association for Computational
                  Linguistics (Volume 1: Long Papers), {ACL} 2024, Bangkok, Thailand,
                  August 11-16, 2024},
  pages        = {6864--6890},
  publisher    = {Association for Computational Linguistics},
  year         = {2024},
  url          = {https://doi.org/10.18653/v1/2024.acl-long.371},
  doi          = {10.18653/V1/2024.ACL-LONG.371},
  bibsource    = {dblp computer science bibliography, https://dblp.org}
}

@inproceedings{Agent-S,
  author       = {Saaket Agashe and
                  Jiuzhou Han and
                  Shuyu Gan and
                  Jiachen Yang and
                  Ang Li and
                  Xin Eric Wang},
  title        = {Agent {S:} An Open Agentic Framework that Uses Computers Like a Human},
  booktitle    = {The Thirteenth International Conference on Learning Representations,
                  {ICLR} 2025, Singapore, April 24-28, 2025},
  publisher    = {OpenReview.net},
  year         = {2025},
  url          = {https://openreview.net/forum?id=lIVRgt4nLv},
  bibsource    = {dblp computer science bibliography, https://dblp.org}
}

@inproceedings{AutoGuide,
  author       = {Yao Fu and
                  Dong{-}Ki Kim and
                  Jaekyeom Kim and
                  Sungryull Sohn and
                  Lajanugen Logeswaran and
                  Kyunghoon Bae and
                  Honglak Lee},
  editor       = {Amir Globersons and
                  Lester Mackey and
                  Danielle Belgrave and
                  Angela Fan and
                  Ulrich Paquet and
                  Jakub M. Tomczak and
                  Cheng Zhang},
  title        = {AutoGuide: Automated Generation and Selection of Context-Aware Guidelines
                  for Large Language Model Agents},
  booktitle    = {Advances in Neural Information Processing Systems 38: Annual Conference
                  on Neural Information Processing Systems 2024, NeurIPS 2024, Vancouver,
                  BC, Canada, December 10 - 15, 2024},
  year         = {2024},
  url          = {http://papers.nips.cc/paper\_files/paper/2024/hash/d8efbb5dd415974eb095c3f06bff1f48-Abstract-Conference.html},
  bibsource    = {dblp computer science bibliography, https://dblp.org}
}

@inproceedings{CRITIC,
  author       = {Zhibin Gou and
                  Zhihong Shao and
                  Yeyun Gong and
                  Yelong Shen and
                  Yujiu Yang and
                  Nan Duan and
                  Weizhu Chen},
  title        = {{CRITIC:} Large Language Models Can Self-Correct with Tool-Interactive
                  Critiquing},
  booktitle    = {The Twelfth International Conference on Learning Representations,
                  {ICLR} 2024, Vienna, Austria, May 7-11, 2024},
  publisher    = {OpenReview.net},
  year         = {2024},
  url          = {https://openreview.net/forum?id=Sx038qxjek},
  bibsource    = {dblp computer science bibliography, https://dblp.org}
}

@inproceedings{SelfRefine,
  author       = {Aman Madaan and
                  Niket Tandon and
                  Prakhar Gupta and
                  Skyler Hallinan and
                  Luyu Gao and
                  Sarah Wiegreffe and
                  Uri Alon and
                  Nouha Dziri and
                  Shrimai Prabhumoye and
                  Yiming Yang and
                  Shashank Gupta and
                  Bodhisattwa Prasad Majumder and
                  Katherine Hermann and
                  Sean Welleck and
                  Amir Yazdanbakhsh and
                  Peter Clark},
  editor       = {Alice Oh and
                  Tristan Naumann and
                  Amir Globerson and
                  Kate Saenko and
                  Moritz Hardt and
                  Sergey Levine},
  title        = {Self-Refine: Iterative Refinement with Self-Feedback},
  booktitle    = {Advances in Neural Information Processing Systems 36: Annual Conference
                  on Neural Information Processing Systems 2023, NeurIPS 2023, New Orleans,
                  LA, USA, December 10 - 16, 2023},
  year         = {2023},
  url          = {http://papers.nips.cc/paper\_files/paper/2023/hash/91edff07232fb1b55a505a9e9f6c0ff3-Abstract-Conference.html},
  bibsource    = {dblp computer science bibliography, https://dblp.org}
}

@inproceedings{AgentLumos,
  author       = {Da Yin and
                  Faeze Brahman and
                  Abhilasha Ravichander and
                  Khyathi Raghavi Chandu and
                  Kai{-}Wei Chang and
                  Yejin Choi and
                  Bill Yuchen Lin},
  editor       = {Lun{-}Wei Ku and
                  Andre Martins and
                  Vivek Srikumar},
  title        = {Agent Lumos: Unified and Modular Training for Open-Source Language
                  Agents},
  booktitle    = {Proceedings of the 62nd Annual Meeting of the Association for Computational
                  Linguistics (Volume 1: Long Papers), {ACL} 2024, Bangkok, Thailand,
                  August 11-16, 2024},
  pages        = {12380--12403},
  publisher    = {Association for Computational Linguistics},
  year         = {2024},
  url          = {https://doi.org/10.18653/v1/2024.acl-long.670},
  doi          = {10.18653/V1/2024.ACL-LONG.670},
  bibsource    = {dblp computer science bibliography, https://dblp.org}
}

@inproceedings{AgentTuning,
  author       = {Aohan Zeng and
                  Mingdao Liu and
                  Rui Lu and
                  Bowen Wang and
                  Xiao Liu and
                  Yuxiao Dong and
                  Jie Tang},
  editor       = {Lun{-}Wei Ku and
                  Andre Martins and
                  Vivek Srikumar},
  title        = {AgentTuning: Enabling Generalized Agent Abilities for LLMs},
  booktitle    = {Findings of the Association for Computational Linguistics, {ACL} 2024,
                  Bangkok, Thailand and virtual meeting, August 11-16, 2024},
  pages        = {3053--3077},
  publisher    = {Association for Computational Linguistics},
  year         = {2024},
  url          = {https://doi.org/10.18653/v1/2024.findings-acl.181},
  doi          = {10.18653/V1/2024.FINDINGS-ACL.181},
  bibsource    = {dblp computer science bibliography, https://dblp.org}
}

@inproceedings{AutoAct,
  author       = {Shuofei Qiao and
                  Ningyu Zhang and
                  Runnan Fang and
                  Yujie Luo and
                  Wangchunshu Zhou and
                  Yuchen Eleanor Jiang and
                  Chengfei Lv and
                  Huajun Chen},
  editor       = {Lun{-}Wei Ku and
                  Andre Martins and
                  Vivek Srikumar},
  title        = {AutoAct: Automatic Agent Learning from Scratch for {QA} via Self-Planning},
  booktitle    = {Proceedings of the 62nd Annual Meeting of the Association for Computational
                  Linguistics (Volume 1: Long Papers), {ACL} 2024, Bangkok, Thailand,
                  August 11-16, 2024},
  pages        = {3003--3021},
  publisher    = {Association for Computational Linguistics},
  year         = {2024},
  url          = {https://doi.org/10.18653/v1/2024.acl-long.165},
  doi          = {10.18653/V1/2024.ACL-LONG.165},
  bibsource    = {dblp computer science bibliography, https://dblp.org}
}

@inproceedings{MACPO,
  author       = {Yougang Lyu and
                  Lingyong Yan and
                  Zihan Wang and
                  Dawei Yin and
                  Pengjie Ren and
                  Maarten de Rijke and
                  Zhaochun Ren},
  title        = {{MACPO:} Weak-to-Strong Alignment via Multi-Agent Contrastive Preference
                  Optimization},
  booktitle    = {The Thirteenth International Conference on Learning Representations,
                  {ICLR} 2025, Singapore, April 24-28, 2025},
  publisher    = {OpenReview.net},
  year         = {2025},
  url          = {https://openreview.net/forum?id=x1Okv4kbVR},
  bibsource    = {dblp computer science bibliography, https://dblp.org}
}

@article{DeepSeekMath,
  author       = {Zhihong Shao and
                  Peiyi Wang and
                  Qihao Zhu and
                  Runxin Xu and
                  Junxiao Song and
                  Mingchuan Zhang and
                  Y. K. Li and
                  Y. Wu and
                  Daya Guo},
  title        = {DeepSeekMath: Pushing the Limits of Mathematical Reasoning in Open
                  Language Models},
  journal      = {CoRR},
  volume       = {abs/2402.03300},
  year         = {2024},
  url          = {https://doi.org/10.48550/arXiv.2402.03300},
  doi          = {10.48550/ARXIV.2402.03300},
  eprinttype    = {arXiv},
  eprint       = {2402.03300},
  bibsource    = {dblp computer science bibliography, https://dblp.org}
}

@article{SPA-RL,
  author       = {Hanlin Wang and
                  Chak Tou Leong and
                  Jiashuo Wang and
                  Jian Wang and
                  Wenjie Li},
  title        = {{SPA-RL:} Reinforcing {LLM} Agents via Stepwise Progress Attribution},
  journal      = {CoRR},
  volume       = {abs/2505.20732},
  year         = {2025},
  url          = {https://doi.org/10.48550/arXiv.2505.20732},
  doi          = {10.48550/ARXIV.2505.20732},
  eprinttype    = {arXiv},
  eprint       = {2505.20732},
  bibsource    = {dblp computer science bibliography, https://dblp.org}
}

@inproceedings{WebVLN,
  author       = {Qi Chen and
                  Dileepa Pitawela and
                  Chongyang Zhao and
                  Gengze Zhou and
                  Hsiang{-}Ting Chen and
                  Qi Wu},
  editor       = {Michael J. Wooldridge and
                  Jennifer G. Dy and
                  Sriraam Natarajan},
  title        = {WebVLN: Vision-and-Language Navigation on Websites},
  booktitle    = {Thirty-Eighth {AAAI} Conference on Artificial Intelligence, {AAAI}
                  2024, Thirty-Sixth Conference on Innovative Applications of Artificial
                  Intelligence, {IAAI} 2024, Fourteenth Symposium on Educational Advances
                  in Artificial Intelligence, {EAAI} 2014, February 20-27, 2024, Vancouver,
                  Canada},
  pages        = {1165--1173},
  publisher    = {{AAAI} Press},
  year         = {2024},
  url          = {https://doi.org/10.1609/aaai.v38i2.27878},
  doi          = {10.1609/AAAI.V38I2.27878},
  bibsource    = {dblp computer science bibliography, https://dblp.org}
}

@inproceedings{WebWalker,
  author       = {Jialong Wu and
                  Wenbiao Yin and
                  Yong Jiang and
                  Zhenglin Wang and
                  Zekun Xi and
                  Runnan Fang and
                  Linhai Zhang and
                  Yulan He and
                  Deyu Zhou and
                  Pengjun Xie and
                  Fei Huang},
  editor       = {Wanxiang Che and
                  Joyce Nabende and
                  Ekaterina Shutova and
                  Mohammad Taher Pilehvar},
  title        = {WebWalker: Benchmarking LLMs in Web Traversal},
  booktitle    = {Proceedings of the 63rd Annual Meeting of the Association for Computational
                  Linguistics (Volume 1: Long Papers), {ACL} 2025, Vienna, Austria,
                  July 27 - August 1, 2025},
  pages        = {10290--10305},
  publisher    = {Association for Computational Linguistics},
  year         = {2025},
  url          = {https://aclanthology.org/2025.acl-long.508/},
  bibsource    = {dblp computer science bibliography, https://dblp.org}
}

@inproceedings{VLNBERT,
  author       = {Yicong Hong and
                  Qi Wu and
                  Yuankai Qi and
                  Cristian Rodriguez Opazo and
                  Stephen Gould},
  title        = {{VLN} {BERT:} {A} Recurrent Vision-and-Language {BERT} for Navigation},
  booktitle    = {{IEEE} Conference on Computer Vision and Pattern Recognition, {CVPR}
                  2021, virtual, June 19-25, 2021},
  pages        = {1643--1653},
  publisher    = {Computer Vision Foundation / {IEEE}},
  year         = {2021},
  url          = {https://openaccess.thecvf.com/content/CVPR2021/html/Hong\_VLN\_BERT\_A\_Recurrent\_Vision-and-Language\_BERT\_for\_Navigation\_CVPR\_2021\_paper.html},
  doi          = {10.1109/CVPR46437.2021.00169},
  bibsource    = {dblp computer science bibliography, https://dblp.org}
}

@inproceedings{LXMERT,
  author       = {Hao Tan and
                  Mohit Bansal},
  editor       = {Kentaro Inui and
                  Jing Jiang and
                  Vincent Ng and
                  Xiaojun Wan},
  title        = {{LXMERT:} Learning Cross-Modality Encoder Representations from Transformers},
  booktitle    = {Proceedings of the 2019 Conference on Empirical Methods in Natural
                  Language Processing and the 9th International Joint Conference on
                  Natural Language Processing, {EMNLP-IJCNLP} 2019, Hong Kong, China,
                  November 3-7, 2019},
  pages        = {5099--5110},
  publisher    = {Association for Computational Linguistics},
  year         = {2019},
  url          = {https://doi.org/10.18653/v1/D19-1514},
  doi          = {10.18653/V1/D19-1514},
  bibsource    = {dblp computer science bibliography, https://dblp.org}
}

@inproceedings{WebGUM,
  author       = {Hiroki Furuta and
                  Kuang{-}Huei Lee and
                  Ofir Nachum and
                  Yutaka Matsuo and
                  Aleksandra Faust and
                  Shixiang Shane Gu and
                  Izzeddin Gur},
  title        = {Multimodal Web Navigation with Instruction-Finetuned Foundation Models},
  booktitle    = {The Twelfth International Conference on Learning Representations,
                  {ICLR} 2024, Vienna, Austria, May 7-11, 2024},
  publisher    = {OpenReview.net},
  year         = {2024},
  url          = {https://openreview.net/forum?id=efFmBWioSc},
  bibsource    = {dblp computer science bibliography, https://dblp.org}
}

@inproceedings{dao2023flashattention,
  author       = {Tri Dao},
  title        = {FlashAttention-2: Faster Attention with Better Parallelism and Work
                  Partitioning},
  booktitle    = {The Twelfth International Conference on Learning Representations,
                  {ICLR} 2024, Vienna, Austria, May 7-11, 2024},
  publisher    = {OpenReview.net},
  year         = {2024},
  url          = {https://openreview.net/forum?id=mZn2Xyh9Ec},
  bibsource    = {dblp computer science bibliography, https://dblp.org}
}

@inproceedings{zhao2024swift,
  author       = {Yuze Zhao and
                  Jintao Huang and
                  Jinghan Hu and
                  Xingjun Wang and
                  Yunlin Mao and
                  Daoze Zhang and
                  Zeyinzi Jiang and
                  Zhikai Wu and
                  Baole Ai and
                  Ang Wang and
                  Wenmeng Zhou and
                  Yingda Chen},
  editor       = {Toby Walsh and
                  Julie Shah and
                  Zico Kolter},
  title        = {{SWIFT:} {A} Scalable Lightweight Infrastructure for Fine-Tuning},
  booktitle    = {AAAI-25, Sponsored by the Association for the Advancement of Artificial
                  Intelligence, February 25 - March 4, 2025, Philadelphia, PA, {USA}},
  pages        = {29733--29735},
  publisher    = {{AAAI} Press},
  year         = {2025},
  url          = {https://doi.org/10.1609/aaai.v39i28.35383},
  doi          = {10.1609/AAAI.V39I28.35383},
  bibsource    = {dblp computer science bibliography, https://dblp.org}
}

@inproceedings{rajbhandari2020zero,
  author       = {Samyam Rajbhandari and
                  Jeff Rasley and
                  Olatunji Ruwase and
                  Yuxiong He},
  editor       = {Christine Cuicchi and
                  Irene Qualters and
                  William T. Kramer},
  title        = {ZeRO: memory optimizations toward training trillion parameter models},
  booktitle    = {Proceedings of the International Conference for High Performance Computing,
                  Networking, Storage and Analysis, {SC} 2020, Virtual Event / Atlanta,
                  Georgia, USA, November 9-19, 2020},
  pages        = {20},
  publisher    = {{IEEE/ACM}},
  year         = {2020},
  url          = {https://doi.org/10.1109/SC41405.2020.00024},
  doi          = {10.1109/SC41405.2020.00024},
  bibsource    = {dblp computer science bibliography, https://dblp.org}
}

@inproceedings{loshchilov2017decoupled,
  author       = {Ilya Loshchilov and
                  Frank Hutter},
  title        = {Decoupled Weight Decay Regularization},
  booktitle    = {7th International Conference on Learning Representations, {ICLR} 2019,
                  New Orleans, LA, USA, May 6-9, 2019},
  publisher    = {OpenReview.net},
  year         = {2019},
  url          = {https://openreview.net/forum?id=Bkg6RiCqY7},
  bibsource    = {dblp computer science bibliography, https://dblp.org}
}

@inproceedings{zhou2024navgpt,
  title={Navgpt: Explicit reasoning in vision-and-language navigation with large language models},
  author={Zhou, Gengze and Hong, Yicong and Wu, Qi},
  booktitle={Proceedings of the AAAI Conference on Artificial Intelligence},
  volume={38},
  number={7},
  pages={7641--7649},
  year={2024},
  url = {https://doi.org/10.1609/aaai.v38i7.28597}
}

@inproceedings{React,
  author       = {Shunyu Yao and
                  Jeffrey Zhao and
                  Dian Yu and
                  Nan Du and
                  Izhak Shafran and
                  Karthik R. Narasimhan and
                  Yuan Cao},
  title        = {ReAct: Synergizing Reasoning and Acting in Language Models},
  booktitle    = {The Eleventh International Conference on Learning Representations,
                  {ICLR} 2023, Kigali, Rwanda, May 1-5, 2023},
  publisher    = {OpenReview.net},
  year         = {2023},
  url          = {https://openreview.net/forum?id=WE\_vluYUL-X},
  bibsource    = {dblp computer science bibliography, https://dblp.org}
}

@inproceedings{YuK0LHF20,
  author       = {Tianhe Yu and
                  Saurabh Kumar and
                  Abhishek Gupta and
                  Sergey Levine and
                  Karol Hausman and
                  Chelsea Finn},
  editor       = {Hugo Larochelle and
                  Marc'Aurelio Ranzato and
                  Raia Hadsell and
                  Maria{-}Florina Balcan and
                  Hsuan{-}Tien Lin},
  title        = {Gradient Surgery for Multi-Task Learning},
  booktitle    = {Advances in Neural Information Processing Systems 33: Annual Conference
                  on Neural Information Processing Systems 2020, NeurIPS 2020, December
                  6-12, 2020, virtual},
  year         = {2020},
  url          = {https://proceedings.neurips.cc/paper/2020/hash/3fe78a8acf5fda99de95303940a2420c-Abstract.html},
  bibsource    = {dblp computer science bibliography, https://dblp.org}
}

@inproceedings{SenerK18,
  author       = {Ozan Sener and
                  Vladlen Koltun},
  editor       = {Samy Bengio and
                  Hanna M. Wallach and
                  Hugo Larochelle and
                  Kristen Grauman and
                  Nicol{\`{o}} Cesa{-}Bianchi and
                  Roman Garnett},
  title        = {Multi-Task Learning as Multi-Objective Optimization},
  booktitle    = {Advances in Neural Information Processing Systems 31: Annual Conference
                  on Neural Information Processing Systems 2018, NeurIPS 2018, December
                  3-8, 2018, Montr{\'{e}}al, Canada},
  pages        = {525--536},
  year         = {2018},
  url          = {https://proceedings.neurips.cc/paper/2018/hash/432aca3a1e345e339f35a30c8f65edce-Abstract.html},
  bibsource    = {dblp computer science bibliography, https://dblp.org}
}

@inproceedings{ShinnCGNY23,
  author       = {Noah Shinn and
                  Federico Cassano and
                  Ashwin Gopinath and
                  Karthik Narasimhan and
                  Shunyu Yao},
  editor       = {Alice Oh and
                  Tristan Naumann and
                  Amir Globerson and
                  Kate Saenko and
                  Moritz Hardt and
                  Sergey Levine},
  title        = {Reflexion: language agents with verbal reinforcement learning},
  booktitle    = {Advances in Neural Information Processing Systems 36: Annual Conference
                  on Neural Information Processing Systems 2023, NeurIPS 2023, New Orleans,
                  LA, USA, December 10 - 16, 2023},
  year         = {2023},
  url          = {http://papers.nips.cc/paper\_files/paper/2023/hash/1b44b878bb782e6954cd888628510e90-Abstract-Conference.html},
  bibsource    = {dblp computer science bibliography, https://dblp.org}
}
